\PassOptionsToPackage{table}{xcolor}
\documentclass{phai}

\ifdefined\mymaketitle
\newcommand{\ccpoShadedRow}{\rowcolor{black!6}[0pt][0pt]}
\newcommand{\ccpoHereFigure}{\noindent\begin{minipage}{\linewidth}}
\newcommand{\ccpoHeadedFigure}[2]{\noindent\begin{minipage}{\linewidth}\section{#1}\label{#2}}
\newcommand{\ccpoHereCaption}{\captionof{figure}}
\newcommand{\ccpoEndHereFigure}{\end{minipage}}
\newcommand{\ccpoMeasure}{\sbox{0}}
\newcommand{\ccpoSubgroupRule}[1]{%
    \noalign{\vskip 2pt}%
    \multispan{#1}%
    \begingroup\color{black!45}%
    \xleaders\hbox{\vrule width 2pt height 0.35pt depth 0pt\hskip 2pt}\hfill
    \endgroup\cr
    \noalign{\vskip 2pt}%
}
\else
\newcommand{\ccpoShadedRow}{}
\newcommand{\ccpoHereFigure}{\begin{figure}}
\newcommand{\ccpoHeadedFigure}[2]{\section{#1}\label{#2}\begin{figure}}
\newcommand{\ccpoHereCaption}{\caption}
\newcommand{\ccpoEndHereFigure}{\end{figure}}
\newcommand{\ccpoSubgroupRule}[1]{\midrule}
\newcommand{\ccpoMeasure}[1]{#1}
\providecommand{\beginappendix}{\appendix}
\fi

\usepackage{amsmath,amsfonts,bm}

\def\eqref#1{equation~\ref{#1}}

\def\1{\bm{1}}

\DeclareMathAlphabet{\mathsfit}{\encodingdefault}{\sfdefault}{m}{sl}
\SetMathAlphabet{\mathsfit}{bold}{\encodingdefault}{\sfdefault}{bx}{n}

\usepackage{url}
\usepackage{array,longtable,tabularx}
\usepackage{fancyvrb}
\usepackage{needspace}
\newcounter{ccpoalgorithm}
\newcommand{\authornewline}[2][]{%
  \addtolist[#1]{#2}{\authorlist}{\authorformat}{, \protect\\}%
}

\newcommand{\ccpoContextMark}[1]{%
    \begingroup
    \setlength{\unitlength}{0.075em}%
    \linethickness{0.035em}%
    \begin{picture}(10,9)
        \put(5,4.5){\circle{8.6}}%
        #1%
    \end{picture}%
    \endgroup
}
\DeclareRobustCommand{\ccpoAnchor}{\ccpoContextMark{\put(5,4.5){\circle*{3.2}}}}
\DeclareRobustCommand{\ccpoHistory}{\ccpoContextMark{\put(5,4.5){\circle*{6.4}}}}
\DeclareRobustCommand{\ccpoFuture}{\ccpoContextMark{\put(5,4.5){\circle{5.2}}}}

\title{Contextual Credit Policy Optimization}

\ifdefined\mymaketitle
\author[1,2,\dagger]{Huiwen Han}
\author[3]{Ziyao Zeng}
\author[2]{Zihao Zhu}
\author[4]{Sisong Bei}
\author[2]{Hao Wang}
\authornewline[5]{Hongyu Xu}
\author[6]{Yan Han}
\author[3]{Degui Zhi}
\author[2]{Zhiwen Fan}

\affiliation[1]{Stanford University}
\affiliation[2]{Texas A\&M University}
\affiliation[3]{Yale University}
\affiliation[4]{Amazon}
\affiliation[5]{Meta AI}
\affiliation[6]{Microsoft}
\contribution[\dagger]{Project lead}

\abstract{Group-based reinforcement learning enables large language model agents to learn from sparse outcome rewards in long-horizon tasks. Recent methods derive step-level credit by comparing observation-matched experiences across trajectories. However, identical observations can occur in different contexts, and matched peers are not equally informative for a given decision. Moreover, favorable terminal outcomes can obscure redundant actions or detours within successful trajectories. To address these two challenges, we propose \textbf{Contextual Credit Policy Optimization (CCPO)}. For the first, CCPO weights observation-matched peers from other trajectories by contextual similarity, computed from frozen LLM representations and accumulated trajectory statistics, allowing peers from similar contexts to contribute more to a step's baseline. For the second, CCPO pairs historical credit, which compares the realized return against the weighted peer baseline, with future credit, which measures the change in contextual outcome value over a short realized continuation. A step is thus credited for the progress it makes, not only for its trajectory's eventual outcome. Both credits are computed in hindsight from rollouts already collected for training, enabling CCPO to address these challenges without an auxiliary value critic. Across two Qwen2.5-Instruct backbones (1.5B and 7B), CCPO achieves up to 97.4\% success on ALFWorld and 79.4\% on WebShop and outperforms GRPO by up to 21.6 and 18.5 points. Ablations show that contextual weighting outperforms uniform weighting, and that historical and future credit serve as complementary learning signals.}
\metadata[Affiliation Note]{Huiwen Han was a visiting student at Texas A\&M University during this work.}
\metadata[Project Page]{\href{https://agent-ccpo.github.io}{\texttt{agent-ccpo.github.io}}}
\metadata[Code]{\href{https://github.com/phai-lab/CCPO}{\texttt{github.com/phai-lab/CCPO}}}
\else
\providecommand{\affiliation}[1]{}
\author{Huiwen Han\textsuperscript{1,2,\textdagger}{,} Ziyao Zeng\textsuperscript{3}{,} Zihao Zhu\textsuperscript{2}{,} Sisong Bei\textsuperscript{4}{,} Hao Wang\textsuperscript{2}{,} Hongyu Xu\textsuperscript{5}{,} Yan Han\textsuperscript{6}{,} Degui Zhi\textsuperscript{3}{,} Zhiwen Fan\textsuperscript{2}}
\affiliation{\textsuperscript{1}Stanford University, \textsuperscript{2}Texas A\&M University, \textsuperscript{3}Yale University, \textsuperscript{4}Amazon, \textsuperscript{5}Meta AI, \textsuperscript{6}Microsoft}
\affiliation{\textsuperscript{\textdagger}Project lead}
\date{}
\fi

\hypersetup{
  pdftitle={Contextual Credit Policy Optimization},
  pdfauthor={Huiwen Han, Ziyao Zeng, Zihao Zhu, Sisong Bei, Hao Wang, Hongyu Xu, Yan Han, Degui Zhi, Zhiwen Fan}
}

\begin{document}
\maketitle
\ifdefined\mymaketitle\else
\begin{abstract}
Group-based reinforcement learning enables large language model agents to learn from sparse outcome rewards in long-horizon tasks. Recent methods derive step-level credit by comparing observation-matched experiences across trajectories. However, identical observations can occur in different contexts, and matched peers are not equally informative for a given decision. Moreover, favorable terminal outcomes can obscure redundant actions or detours within successful trajectories. To address these two challenges, we propose \textbf{Contextual Credit Policy Optimization (CCPO)}. For the first, CCPO weights observation-matched peers from other trajectories by contextual similarity, computed from frozen LLM representations and accumulated trajectory statistics, allowing peers from similar contexts to contribute more to a step's baseline. For the second, CCPO pairs historical credit, which compares the realized return against the weighted peer baseline, with future credit, which measures the change in contextual outcome value over a short realized continuation. A step is thus credited for the progress it makes, not only for its trajectory's eventual outcome. Both credits are computed in hindsight from rollouts already collected for training, enabling CCPO to address these challenges without an auxiliary value critic. Across two Qwen2.5-Instruct backbones (1.5B and 7B), CCPO achieves up to 97.4\% success on ALFWorld and 79.4\% on WebShop and outperforms GRPO by up to 21.6 and 18.5 points. Ablations show that contextual weighting outperforms uniform weighting, and that historical and future credit serve as complementary learning signals.
\end{abstract}

\noindent\textbf{Affiliation Note:} Huiwen Han was a visiting student at Texas A\&M University during this work.\\
\textbf{Project Page:} \url{https://agent-ccpo.github.io}\\
\textbf{Code:} \url{https://github.com/phai-lab/CCPO}
\fi

\begin{figure}[!ht]
    \centering
    \vspace{-4pt}
    \includegraphics[width=\linewidth]{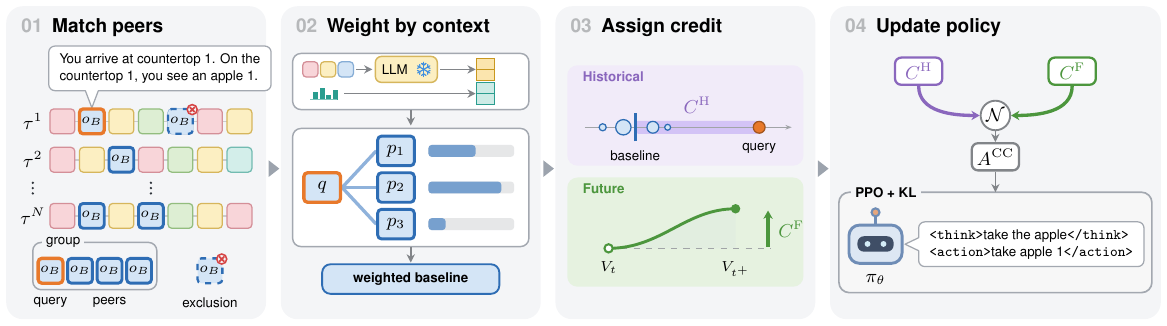}
    \caption{\textbf{Overview of CCPO.} CCPO combines contextually weighted peer comparisons and short-horizon future progress for critic-free step-level credit assignment.}
    \label{fig:ccpo-teaser}
    \vspace{-4pt}
\end{figure}

\section{Introduction}
Large language model (LLM) agents increasingly undertake long-horizon tasks through sustained environment interaction~\citep{wang2025surveylargelanguagemodel,zhang2026landscapeagenticreinforcementlearning}. Benchmarks such as WebArena~\citep{zhou2024webarenarealisticwebenvironment} and SWE-bench~\citep{jimenez2024swebenchlanguagemodelsresolve} require information gathering, coordinated actions, and adaptation to feedback. Reinforcement learning (RL)~\citep{sutton2018reinforcement} provides a post-training framework that optimizes policies using rewards from interaction trajectories. Recent applications span open-web research, web navigation, and multi-step software engineering~\citep{zheng-etal-2025-deepresearcher,qi2025webrltrainingllmweb,deepswe2025}. However, feedback often arrives only after many decisions.

Learning from such delayed feedback depends on how it is translated into policy updates. PPO-based actor--critic methods use learned value estimates to guide optimization, at the cost of training an additional value model~\citep{schulman2017proximalpolicyoptimizationalgorithms}. Alternatively, group-based methods such as GRPO~\citep{shao2024deepseekmathpushinglimitsmathematical} estimate relative advantages from multiple sampled outcomes without a separately learned value critic. When applied at the trajectory level, however, a single advantage is shared across all actions in a rollout. Such comparisons differentiate outcomes across trajectories but provide limited information about the contributions of individual decisions, especially when rewards are sparse and delayed~\citep{zhang2026landscapeagenticreinforcementlearning}. To enable fine-grained policy updates, recent methods extend this approach to construct step-level advantage estimates: GiGPO~\citep{feng2025groupingrouppolicyoptimizationllm} compares actions encountered at shared observations, and G2PO~\citep{wang2026groupgraphpolicyoptimizationlonghorizon} incorporates the structure of transitions between observation groups. HGPO~\citep{he2026hierarchyofgroupspolicyoptimizationlonghorizon} accounts for historical context by organizing comparisons into hierarchical groups. In each case, step-level credit comes from comparing experiences in existing rollouts, so it depends on how they are compared.

Two challenges remain when steps are compared at matched observations. The first is whom a step should be compared with. Identical observations can occur after different preceding actions, reasoning, exploration depth, and accumulated contexts. Peers that share an observation are therefore not equally informative for the step being evaluated (the query). Requiring peers to share the same history instead leaves few peers per step, and a baseline averaged over so few returns gives noisy advantages~\citep{he2026hierarchyofgroupspolicyoptimizationlonghorizon}. Our premise is that, within observation-matched groups, contextual similarity should determine each peer's contribution to the query's baseline and value estimates. The second is what a step should be credited for. A return comparison credits a step for belonging to a trajectory that ended well, but an eventual success can include redundant actions or detours. Separating a step that inherits a favorable context from one that makes progress from it requires a complementary comparison of realized gains over a short continuation.

\looseness=-1 To address both challenges, we propose \textbf{Contextual Credit Policy Optimization (CCPO)}. For the first, CCPO represents each step's context with semantic features of an interaction window from a frozen LLM and accumulated trajectory statistics, and gives greater weight to observation-matched peers with higher contextual similarity (Figure~\ref{fig:ccpo-teaser}). For the second, CCPO builds two credit signals on these weighted peers. Historical credit compares the realized return with its contextual peer baseline. Future credit measures short-horizon progress as the change in contextual value, a weighted average of peers' discounted terminal rewards, between the current step and a later step of the same rollout. Both are computed in hindsight from collected rollouts without requiring an auxiliary value critic or additional environment interactions. On agentic multi-step benchmarks, ALFWorld~\citep{shridhar2021alfworldaligningtextembodied} and WebShop~\citep{yao2023webshopscalablerealworldweb}, CCPO outperforms prompting- and RL-based baselines using Qwen2.5-1.5B-Instruct and Qwen2.5-7B-Instruct backbones~\citep{qwen2025qwen25technicalreport}.

Our main contributions are as follows:
\begin{itemize}
    \item \textbf{Contextual credit estimation from history and future progress.}
    We assign step-level credit by combining a history-conditioned return residual with estimated value change over a future horizon, capturing outcome-relative performance and local progress. Analysis shows complementary effects of history and future credit as learning evolves.

    \item \textbf{Contextual similarity for step-level peer weighting.}
    We aggregate peer returns with continuous weights set by the similarity between query and peer context representations, producing a context-dependent baseline. Replacing these weights with uniform ones lowers success by 3.7 and 2.6 points.

    \item \textbf{Semantic and structural context representations.}
    We combine semantic hidden states of recent interactions with accumulated trajectory statistics, enabling context-sensitive comparisons across trajectories without training an auxiliary value critic.
\end{itemize}

\section{Related Work}
\paragraph{Agentic reinforcement learning for long-horizon tasks.} Reinforcement learning (RL)~\citep{sutton2018reinforcement} enables large language model (LLM) agents to improve through extended interactions in which observations, reasoning, actions, and feedback accumulate into trajectories~\citep{pmlr-v235-zhou24t,zhang2026landscapeagenticreinforcementlearning}. Applications span interactive gaming, web navigation, open-web research, and multi-step software engineering~\citep{wang2025ragenunderstandingselfevolutionllm,qi2025webrltrainingllmweb,zheng-etal-2025-deepresearcher,qu2026literesearcher,deepswe2025}, where agents act in an environment and adapt to its feedback over many steps. Another line of work trains agents to reason with external tools, including search engines, web browsing tools, and code interpreters~\citep{jin2025searchr1trainingllmsreason,tongyi2026deepresearch,feng2025retoolreinforcementlearningstrategic}. Agentic RL can also learn from successes and failures through verbal reflection and trajectory preference learning~\citep{shinn2023reflexionlanguageagentsverbal,yao2024retroformerretrospectivelargelanguage,song-etal-2024-trial,putta2024agentqadvancedreasoning}. To collect high-quality trajectories, methods use uncertainty-guided branching and resampling~\citep{dong2025agenticreinforcedpolicyoptimization,wang2026t2pouncertaintyguidedexplorationcontrol}. Reward design~\citep{wang2026ragen2reasoningcollapseagentic} is also crucial, spanning outcome-supervised reward models~\citep{qi2025webrltrainingllmweb}, process rewards from environment-aware models such as DataPRM~\citep{qiu2026rewarding}, and turn-level reward assignment~\citep{tao2026traceturnlevelrewardassignment}. Despite these advances, sparse and delayed rewards, costly environment interaction, and training instability remain challenges~\citep{tao2026traceturnlevelrewardassignment,qu2026literesearcher}, motivating efficient and stable policy optimization from trajectories already collected.

\vspace{-1pt}
\paragraph{Policy optimization for LLM agents.} \looseness=-1 Existing approaches differ in their optimization objectives and advantage estimators. Preference-based methods, including DPO~\citep{rafailov2024directpreferenceoptimizationlanguage}, ETO~\citep{song-etal-2024-trial}, and Agent Q~\citep{putta2024agentqadvancedreasoning}, optimize policies from response or trajectory preferences, including successful and failed agent rollouts, without explicit reward-model fitting. Actor--critic methods use learned value estimates, as in PPO-based reinforcement learning from human feedback (RLHF) with InstructGPT~\citep{ouyang2022traininglanguagemodelsfollow} and multi-step agent learning with ArCHer~\citep{pmlr-v235-zhou24t}. To avoid a separate value critic, group-based methods estimate advantages from multiple sampled responses or trajectories. RLOO~\citep{ahmadian2024basicsrevisitingreinforcestyle} uses the mean reward of other samples as a baseline, while GRPO~\citep{shao2024deepseekmathpushinglimitsmathematical}, used in DeepSeek-R1~\citep{deepseekai2026deepseekr1incentivizingreasoningcapability}, standardizes group rewards. Related variants include LOOP~\citep{chen2025reinforcementlearninglonghorizoninteractive} with leave-one-out PPO updates and Dr.~GRPO~\citep{liu2025understandingr1zeroliketrainingcritical} with revised length and reward normalization. Further refinements target update stability and exploration efficiency through clipping, importance weighting, and sampling, including DAPO~\citep{yu2025dapoopensourcellmreinforcement}, GSPO~\citep{zheng2025groupsequencepolicyoptimization}, CISPO~\citep{minimax2025minimaxm1scalingtesttimecompute}, SAPO~\citep{gao2025softadaptivepolicyoptimization}, VESPO~\citep{shen2026vespovariationalsequencelevelsoft}, SAMPO~\citep{wang2026arlarena}, Clip-Cov~\citep{cui2025entropymechanismreinforcementlearning}, and T$^2$PO~\citep{wang2026t2pouncertaintyguidedexplorationcontrol}. However, trajectory-level applications of outcome-supervised GRPO share a single advantage across all actions in a rollout, without explicitly differentiating individual contributions~\citep{feng2025groupingrouppolicyoptimizationllm}. This motivates the step-level methods discussed next.

\vspace{-1pt}
\paragraph{Step-level credit assignment.} Recent methods construct step-level learning signals for multi-step interactions~\citep{zhang2026reasoningagentic}. Critic-based methods include SWEET-RL~\citep{zhou2025sweetrltrainingmultiturnllm} with additional training-time information and HCAPO~\citep{tan2026hindsightcreditassignmentlonghorizon} with the agent as a hindsight critic. TRACE~\citep{tao2026traceturnlevelrewardassignment} derives turn-level rewards from a frozen model's gold-answer likelihoods. OpenClaw-RL~\citep{wang2026openclawrl} combines evaluative rewards and corrective hints extracted from next-state feedback. Other methods estimate advantages by comparing collected trajectories. GiGPO~\citep{feng2025groupingrouppolicyoptimizationllm} combines episode-level advantages with comparisons at matching observations, while HGPO~\citep{he2026hierarchyofgroupspolicyoptimizationlonghorizon} aggregates advantages from hierarchical groups matched at different history lengths. SALT~\citep{li-etal-2026-salt} constructs graphs from trajectories of the same prompt, while G2PO~\citep{wang2026groupgraphpolicyoptimizationlonghorizon} aggregates observations into a state-transition graph to estimate values and advantages. GEPO~\citep{yuan2026graphenhancedpolicyoptimizationllm} uses task-conditioned graph centrality to shape trajectory returns and scale step-level advantages. Peer matching faces a granularity trade-off: observation-only matching can mix different histories, whereas exact history matching reduces the available samples~\citep{he2026hierarchyofgroupspolicyoptimizationlonghorizon}. CCPO keeps all observation-matched peers and weights each by the similarity of its context to the query's, characterized by local semantic and accumulated trajectory features, thus enabling fine-grained comparisons utilizing available peers.

\section{Contextual Credit Policy Optimization}
\label{sec:method}

\begin{figure}[t]
    \centering
    \vspace{-4pt}
    \includegraphics[width=\linewidth]{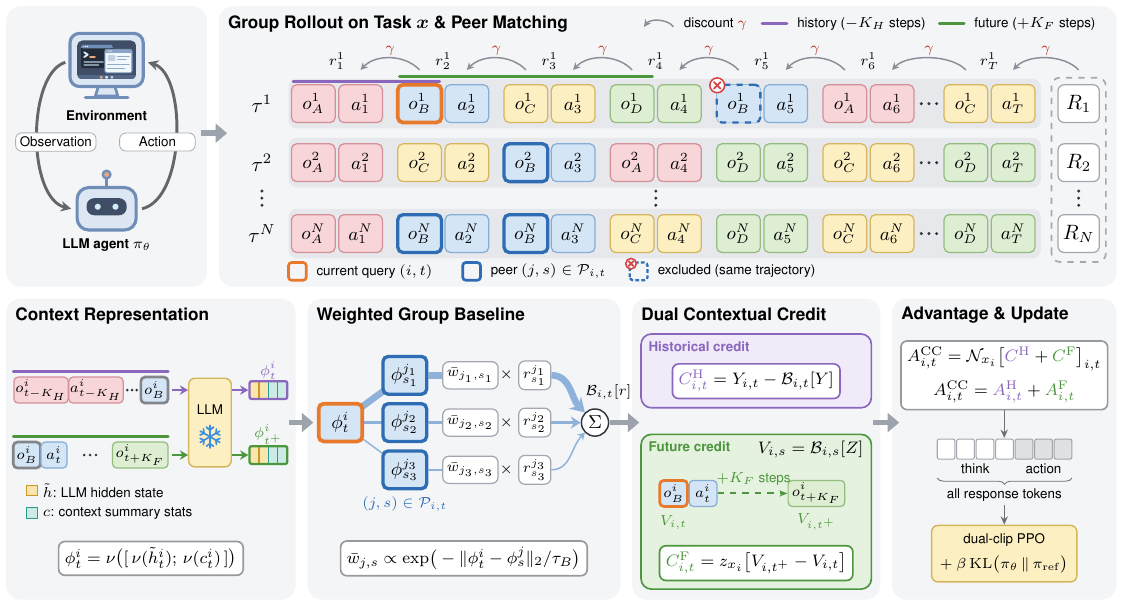}
    \caption{\textbf{Contextual credit policy optimization of multi-step tasks.} Contextual similarity weights observation-matched peers. A history-conditioned return residual and estimated outcome-value change over a short future window yield step-level credit from collected rollouts without an auxiliary value critic.}
    \label{fig:ccpo-overview}
    \vspace{-4pt}
\end{figure}

\subsection{Problem Formulation}
\label{sec:method-setup}
Notation is summarized in Appendix~\ref{s01-notation}.
We sample $N$ trajectories per task from behavior policy $\pi_{\mathrm{old}}$. Trajectory $i$ for task $x_i$ contains $T_i$ actions $a_{i,0:T_i-1}$, observations $o_{i,0:T_i}$, and terminal outcome $R_i$. Prompt $p_{i,t}$ contains $x_i$, $o_{i,t}$, and up to $K_{\mathrm H}$ preceding observation--action pairs. Rewards come from environment feedback. GRPO~\citep{shao2024deepseekmathpushinglimitsmathematical} uses the mean reward of group rollouts as a baseline to estimate normalized advantages without a learned value critic. CCPO extends this critic-free estimation to individual steps, weighting peers by contextual similarity to estimate return baselines. The return target used by historical credit is
\begin{equation}
    Y_{i,t}=\sum_{u=t}^{T_i-1}\gamma^{u-t}r^{\mathrm{env}}_{i,u}
    -\eta\,\mathbf{1}\{a_{i,t}\text{ is invalid}\}.
    \label{eq:historical-target}
\end{equation}
Here $r^{\mathrm{env}}_{i,u}$ follows action $a_{i,u}$ and $\gamma$ discounts rewards. The penalty term subtracts $\eta$ for an invalid action at step $t$ only (Appendix~\ref{app:ccpo-normalization}).

\subsection{Contextual Peer Estimation}
\label{sec:contextual-peer-estimation}
CCPO assigns continuous, query-specific weights to individual peers based on contextual similarity (Figure~\ref{fig:ccpo-overview}). Averaging uniformly within an observation-matched group ignores that identical observations can follow different histories. Soft weighting within observation-matched groups accounts for these differences without exact history matching, which can leave few peers~\citep{he2026hierarchyofgroupspolicyoptimizationlonghorizon}.

\paragraph{Peer selection.}
Exact peers share the task and observation and exclude the entire trajectory of the query step:
\begin{equation}
    \mathcal E_{i,t}
    =\{(j,s):x_j=x_i,\ j\neq i,\ 0\leq s<T_j,\ o_{j,s}=o_{i,t}\}.
    \label{eq:exact-peer-set}
\end{equation}
We use $\mathcal P_{i,t}=\mathcal E_{i,t}$ when nonempty; otherwise, we use all same-task steps with $j\ne i$. We quantify fallback rates and peer availability during training in Appendix~\ref{app:exact-match-fallback}.

\paragraph{Context representation.}
Each step combines a frozen LLM representation $h_{i,t}$ with a trajectory summary $c_{i,t}$. The representation is the final-layer hidden state of the last prompt token. The summary records elapsed steps, distinct-observation count, novelty fraction, and a revisit indicator through step $t$. Because the prompt holds only the most recent steps, the hidden state cannot reflect earlier history, and the summary supplies it. We L2-normalize each preprocessed component, concatenate them with equal weights, and L2-normalize to obtain $\phi_{i,t}$.

\paragraph{Contextual readout.}
Bucket $B$ contains all batch steps matching the query's task and observation, or task alone for fallback. We aggregate scalar targets $r$ over nonempty $\mathcal P_{i,t}$:
\begin{equation}
    \bar w_{(i,t),(j,s)}
    \propto \exp\bigl(-\|\phi_{i,t}-\phi_{j,s}\|_2/\tau_B\bigr),
    \qquad
    \sum_{(j,s)\in\mathcal P_{i,t}}
    \bar w_{(i,t),(j,s)}=1.
    \label{eq:contextual-kernel}
\end{equation}
\begin{equation}
    \mathcal B_{i,t}[r]
    =\sum_{(j,s)\in\mathcal P_{i,t}}
    \bar w_{(i,t),(j,s)}r_{j,s}.
    \label{eq:contextual-estimator}
\end{equation}
The bandwidth $\tau_B$ is proportional to the median pairwise context distance in $B$, computed before trajectory exclusion. This readout estimates both return baselines and outcome values.

\subsection{Temporal Credit Estimation}
\label{sec:temporal-credit}
CCPO conditions return comparisons on a $K_{\mathrm H}$-step history window and assesses progress over a short realized continuation of $K_{\mathrm F}$ steps. The future window compares discounted contextual outcome values at its endpoints. Historical credit depends on how the step's trajectory ends, whereas future credit asks whether the step itself moved the trajectory to a more valuable context.

\paragraph{Historical credit.}
\label{sec:history-credit}
We subtract the contextual peer baseline from the return:
\begin{equation}
    C^{\mathrm H}_{i,t}=Y_{i,t}-\mathcal B_{i,t}[Y].
    \label{eq:history-credit}
\end{equation}

\paragraph{Future credit.}
\label{sec:future-credit}
We estimate contextual values from unpenalized terminal outcomes:
\begin{equation}
    Z_{i,s}=\gamma^{T_i-s}R_i,
    \qquad
    V_{i,s}=\begin{cases}
        \mathcal B_{i,s}[Z], & s<T_i,\\
        R_i, & s=T_i.
    \end{cases}
    \label{eq:contextual-value}
\end{equation}
\label{eq:outcome-target}
For future horizon $K_{\mathrm F}$, let $t^+=\min(t+K_{\mathrm F},T_i)$. We compute
\begin{equation}
    \Delta V_{i,t}=V_{i,t^+}-V_{i,t},
    \qquad C^{\mathrm F}_{i,t}=z_{x_i}[\Delta V]_{i,t},
    \label{eq:future-credit}
\end{equation}
where $z_x$ standardizes eligible differences within task $x$. Each nonterminal endpoint uses its own context and peer set, and future endpoints are used only during training. The endpoint difference measures growth in discounted contextual value, including changes due to shorter remaining time to success. Compared with a discount-aligned TD residual~\citep{schulman2018highdimensionalcontinuouscontrolusing}, this difference includes an extra nonnegative term that grows with the endpoint value and can survive task-wise centering, so future credit also rewards being closer to the end of a successful trajectory (Appendix~\ref{app:ccpo-normalization}).

\subsection{Policy Optimization}
\label{sec:method-optimization}
We add historical credit and task-standardized future credit with unit coefficients:
\begin{equation}
    A^{\mathrm{CC}}_{i,t}
    =\mathcal N_{x_i}[C^{\mathrm H}+C^{\mathrm F}]_{i,t}.
    \label{eq:ccpo-advantage}
\end{equation}
The normalized contributions of the two credits, defined in Appendix~\ref{app:ccpo-normalization}, satisfy $A^{\mathrm{CC}}=A^{\mathrm H}+A^{\mathrm F}$. The default uses $A_{i,t}=A^{\mathrm{CC}}_{i,t}$. To test whether trajectory-level signals complement step-level credit, the episode-augmented variant adds a GRPO-style episode advantage~\citep{shao2024deepseekmathpushinglimitsmathematical}, $A_{i,t}=A^{\mathrm{CC}}_{i,t}+A^{\mathrm{EP}}_{i,t}$, with unit weight and no further normalization. Each step's advantage weights response tokens in a dual-clipped objective~\citep{ye2020masteringcomplexcontrolmoba}, with a KL penalty toward the frozen reference policy. See Appendix~\ref{app:ccpo-implementation} for details.

\section{Experiments}

\subsection{Experimental Setup}
\paragraph{Benchmarks and metrics.} We evaluate CCPO against baseline methods on ALFWorld~\citep{shridhar2021alfworldaligningtextembodied}
and WebShop~\citep{yao2023webshopscalablerealworldweb}, covering household
task execution and online shopping through multi-step interaction.
ALFWorld comprises six household task categories, with 3,553 training
tasks and 140 held-out tasks of the in-distribution evaluation
split. For WebShop, we follow the goal partition implemented in
verl-agent~\citep{feng2025groupingrouppolicyoptimizationllm},
reserving the first 500 goals as the evaluation pool and sampling
training tasks from the remaining goals. Following prior work~\citep{feng2025groupingrouppolicyoptimizationllm,wang2026groupgraphpolicyoptimizationlonghorizon}, we report overall and per-category success rates on ALFWorld. For WebShop, we report success rate and mean task score.

\paragraph{Models and baselines.} We use Qwen2.5-1.5B-Instruct and Qwen2.5-7B-Instruct~\citep{qwen2025qwen25technicalreport}
as policy backbones. We compare CCPO with three categories of baselines:
(1) \textbf{Off-the-shelf LLMs:}
Gemini 3.8 Flash~\citep{googledeepmind2026gemini38flash},
DeepSeek-V4.1-Flash~\citep{deepseekai2026deepseekv41flashpushinglimitskv},
and Qwen3.8-27B~\citep{qwen38}.
(2) \textbf{Prompt-based agents:}
the Qwen2.5-Instruct backbones without task-specific RL fine-tuning,
ReAct~\citep{yao2023reactsynergizingreasoningacting},
and Reflexion~\citep{shinn2023reflexionlanguageagentsverbal}.
(3) \textbf{RL training methods:}
PPO~\citep{schulman2017proximalpolicyoptimizationalgorithms},
RLOO~\citep{ahmadian2024basicsrevisitingreinforcestyle},
GRPO~\citep{shao2024deepseekmathpushinglimitsmathematical},
GiGPO~\citep{feng2025groupingrouppolicyoptimizationllm},
HGPO~\citep{he2026hierarchyofgroupspolicyoptimizationlonghorizon},
and G2PO~\citep{wang2026groupgraphpolicyoptimizationlonghorizon}.
Baseline configurations are detailed in Appendix~\ref{app:training-config}.

\paragraph{Implementation details.} \looseness=-1 We train CCPO using verl-agent~\citep{feng2025groupingrouppolicyoptimizationllm},
built on veRL~\citep{sheng2024hybridflow}, for 150 iterations, sampling 16 tasks with eight trajectories each per iteration. We use a learning rate of $10^{-6}$ and discount factor 0.95. Interaction limits are 50 steps for ALFWorld and 15 for WebShop. Settings are shared across both backbones and all group-based baselines, except that HGPO retains its published limits of 4,096 prompt tokens on ALFWorld and 30 interaction steps on WebShop. CCPO uses a $K_H=2$ history window and a $K_F = 2$ future horizon. We evaluate two CCPO variants, a default and an episode-augmented one. The default uses $A_{\mathrm{CC}}$ from Equation~(\ref{eq:ccpo-advantage}). The variant uses $A_{\mathrm{CC}}+A_{\mathrm{EP}}$, adding a standard episode-level advantage without further normalization. All training and evaluation are completed on 8 $\times$ H200 GPUs. We report means and standard deviations across three random seeds. Appendix~\ref{app:training-config} gives the shared training budgets and optimization settings. Appendix~\ref{app:rollout-examples} provides an ALFWorld prompt and response example.

\subsection{Main Results}

\paragraph{Comparison with baselines.}
\looseness=-1
Across CCPO variants, mean success gains over GRPO are 16.6--21.6 percentage points on ALFWorld and 12.3--18.5 on WebShop at 1.5B. At 7B, they are 12.3--15.4 and 5.5--6.5 points (Table~\ref{tab:main-results}). WebShop score gains are 13.1--14.8 at 1.5B and 1.7--6.5 at 7B. CCPO leads the listed RL baselines in mean success, reaching 96.9\% and 97.4\% on ALFWorld and 75.0\% and 79.4\% on WebShop at 1.5B and 7B.

\begin{table}[t]
\vspace{-4pt}
\centering
\begingroup
\definecolor{ccpoMetricGray}{HTML}{f0f0f0}
\definecolor{ccpoBestBlue}{HTML}{b9daf6}
\definecolor{ccpoReportedGray}{HTML}{999999}
\newcommand{\ccpoReported}[1]{\textcolor{ccpoReportedGray}{#1}}
\setlength{\fboxsep}{1.5pt}
\caption{\textbf{Results on WebShop and ALFWorld.} We report the success rate (\%)
for each sub-task and overall results for ALFWorld, and the success
rate (\%) and score for WebShop, averaged over 3 seeds with standard deviations. Within each section, the \protect\colorbox{ccpoBestBlue}{\textbf{best}} means are highlighted and the \underline{second best} are underlined. Higher is better. \textsuperscript{*}~marks baselines reproduced
or off-the-shelf models evaluated by us. Light-gray rows show results from the
original papers and are excluded from ranking highlights. $\Delta$ GRPO gives the min--max gain over GRPO across both CCPO variants.}
\label{tab:main-results}
\setlength{\tabcolsep}{3pt}
\renewcommand{\arraystretch}{1.15}
\small
\resizebox{\linewidth}{!}{%
\begin{tabular}{@{}ll>{\columncolor{ccpoMetricGray}}c!{\hspace{2pt}{\color{black!35}\vrule width 0.35pt}\hspace{2pt}}*{6}{c}|>{\columncolor{ccpoMetricGray}}c>{\columncolor{ccpoMetricGray}}c}
\toprule
\textbf{Type} & \textbf{Method} & \multicolumn{7}{c|}{\textbf{ALFWorld}} & \multicolumn{2}{c}{\textbf{WebShop}} \\
& & \textbf{All} & Pick & Look & Clean & Heat & Cool & Pick2 & Succ. & Score \\
\midrule
\multicolumn{11}{@{}l}{\textit{Off-the-Shelf Model}} \\
Prompting & Gemini 3.8 Flash\textsuperscript{*}
    & \cellcolor{ccpoBestBlue}\(\mathbf{96.4}\) & \cellcolor{ccpoBestBlue}\(\mathbf{100.0}\) & \cellcolor{ccpoBestBlue}\(\mathbf{84.6}\) & \cellcolor{ccpoBestBlue}\(\mathbf{100.0}\) & \cellcolor{ccpoBestBlue}\(\mathbf{100.0}\) & \cellcolor{ccpoBestBlue}\(\mathbf{92.0}\) & \cellcolor{ccpoBestBlue}\(\mathbf{95.8}\) & \cellcolor{ccpoBestBlue}\(\mathbf{49.2}\) & \cellcolor{ccpoBestBlue}\(\mathbf{54.0}\) \\
Prompting & DeepSeek-V4.1-Flash\textsuperscript{*}
    & \(\underline{75.0}\) & \(\underline{88.6}\) & \cellcolor{ccpoBestBlue}\(\mathbf{84.6}\) & \(\underline{66.7}\) & \(\underline{62.5}\) & \(\underline{76.0}\) & \(\underline{66.7}\) & \(22.0\) & \(27.9\) \\
Prompting & Qwen3.8-27B\textsuperscript{*}
    & \(70.7\) & \(85.7\) & \(\underline{61.5}\) & \(\underline{66.7}\) & \(56.3\) & \(72.0\) & \(\underline{66.7}\) & \(\underline{29.0}\) & \(\underline{39.5}\) \\
\midrule
\multicolumn{11}{@{}l}{\textit{Qwen2.5-1.5B-Instruct}} \\
Prompting & Qwen2.5\textsuperscript{*}
    & \(6.4\) & \(11.4\) & \(7.7\) & \(3.7\) & \(6.3\) & \(4.0\) & \(4.2\) & \(5.0\) & \(21.9\) \\
Prompting & ReAct
    & \(12.8\) & \(17.4\) & \(20.5\) & \(15.7\) & \(6.2\) & \(7.7\) & \(2.0\) & \(11.3\) & \(40.1\) \\
Prompting & Reflexion
    & \(21.8\) & \(35.3\) & \(22.2\) & \(21.7\) & \(13.6\) & \(19.4\) & \(3.7\) & \(21.9\) & \(55.8\) \\
RL Training & PPO (with critic)
    & \(54.4_{\pm 3.1}\) & \(64.8_{\pm 3.5}\) & \(40.5_{\pm 6.9}\) & \(57.1_{\pm 4.9}\) & \(60.6_{\pm 6.6}\) & \(46.4_{\pm 4.0}\) & \(47.4_{\pm 1.9}\) & \(51.5_{\pm 2.9}\) & \(73.8_{\pm 3.0}\) \\
RL Training & RLOO
    & \(69.7_{\pm 2.5}\) & \(88.3_{\pm 3.0}\) & \(52.8_{\pm 8.6}\) & \(71.0_{\pm 5.9}\) & \(62.8_{\pm 8.7}\) & \(66.4_{\pm 5.5}\) & \(56.9_{\pm 4.7}\) & \(52.1_{\pm 6.7}\) & \(73.9_{\pm 5.6}\) \\
RL Training & GRPO\textsuperscript{*}
    & \(75.3_{\pm 1.0}\) & \(81.8_{\pm 0.8}\) & \(66.5_{\pm 8.7}\) & \(77.2_{\pm 0.4}\) & \(73.8_{\pm 3.4}\) & \(78.0_{\pm 3.0}\) & \(65.7_{\pm 1.9}\) & \(56.5_{\pm 2.9}\) & \(70.5_{\pm 2.2}\) \\
RL Training & GiGPO\textsuperscript{*}
    & \(83.9_{\pm 0.7}\) & \(84.8_{\pm 0.7}\) & \(77.0_{\pm 10.0}\) & \(92.4_{\pm 0.1}\) & \cellcolor{ccpoBestBlue}\(\mathbf{100.0}_{\pm 0.0}\) & \(75.0_{\pm 0.9}\) & \(74.6_{\pm 1.9}\) & \(67.4_{\pm 2.2}\) & \(80.1_{\pm 1.8}\) \\
\ccpoReported{RL Training} & \ccpoReported{GiGPO}
    & \ccpoReported{\(86.7_{\pm 1.7}\)} & \ccpoReported{\(94.4_{\pm 5.9}\)} & \ccpoReported{\(67.5_{\pm 4.6}\)} & \ccpoReported{\(94.8_{\pm 3.8}\)} & \ccpoReported{\(94.4_{\pm 7.8}\)} & \ccpoReported{\(79.8_{\pm 4.7}\)} & \ccpoReported{\(76.4_{\pm 5.4}\)} & \ccpoReported{\(65.0_{\pm 3.2}\)} & \ccpoReported{\(83.1_{\pm 1.6}\)} \\
RL Training & HGPO\textsuperscript{*}
    & \(89.1_{\pm 0.6}\) & \(\underline{93.9}_{\pm 0.3}\) & \(80.1_{\pm 6.0}\) & \(92.4_{\pm 0.1}\) & \cellcolor{ccpoBestBlue}\(\mathbf{100.0}_{\pm 0.0}\) & \(88.3_{\pm 1.6}\) & \(76.2_{\pm 1.6}\) & \(65.4_{\pm 1.5}\) & \(82.1_{\pm 0.8}\) \\
\ccpoReported{RL Training} & \ccpoReported{HGPO}
    & \ccpoReported{\(92.8_{\pm 1.1}\)} & \multicolumn{6}{c|}{\ccpoReported{Unreported}} & \ccpoReported{\(71.5_{\pm 4.0}\)} & \ccpoReported{\(85.6_{\pm 2.9}\)} \\
RL Training & G2PO\textsuperscript{*}
    & \(\underline{94.0}_{\pm 0.7}\) & \(\underline{93.9}_{\pm 0.3}\) & \(\underline{90.6}_{\pm 7.4}\) & \cellcolor{ccpoBestBlue}\(\mathbf{100.0}_{\pm 0.0}\) & \cellcolor{ccpoBestBlue}\(\mathbf{100.0}_{\pm 0.0}\) & \(\underline{92.7}_{\pm 1.8}\) & \(\underline{86.6}_{\pm 0.3}\) & \(\underline{71.6}_{\pm 0.7}\) & \(83.5_{\pm 0.8}\) \\
\ccpoReported{RL Training} & \ccpoReported{G2PO}
    & \ccpoReported{\(95.0_{\pm 0.8}\)} & \ccpoReported{\(97.2_{\pm 3.9}\)} & \ccpoReported{\(96.3_{\pm 5.2}\)} & \ccpoReported{\(95.1_{\pm 1.8}\)} & \ccpoReported{\(94.9_{\pm 3.6}\)} & \ccpoReported{\(97.1_{\pm 2.0}\)} & \ccpoReported{\(91.4_{\pm 4.1}\)} & \ccpoReported{\(71.2_{\pm 2.6}\)} & \ccpoReported{\(85.1_{\pm 1.4}\)} \\
\ccpoSubgroupRule{11}
RL Training & \textbf{CCPO (ours)} & \multicolumn{9}{c}{} \\
& \(A_{\mathrm{CC}}\) only
    & \cellcolor{ccpoBestBlue}\(\mathbf{96.9}_{\pm 0.8}\) & \cellcolor{ccpoBestBlue}\(\mathbf{100.0}_{\pm 0.0}\) & \(75.2_{\pm 4.5}\) & \cellcolor{ccpoBestBlue}\(\mathbf{100.0}_{\pm 0.0}\) & \(\underline{93.1}_{\pm 0.7}\) & \cellcolor{ccpoBestBlue}\(\mathbf{100.0}_{\pm 0.0}\) & \cellcolor{ccpoBestBlue}\(\mathbf{97.0}_{\pm 2.6}\) & \(68.8_{\pm 3.4}\) & \(\underline{83.6}_{\pm 2.6}\) \\
\addlinespace[2pt]
& \(A_{\mathrm{CC}} + A_{\mathrm{EP}}\)
    & \(91.9_{\pm 0.5}\) & \(\underline{93.9}_{\pm 0.2}\) & \cellcolor{ccpoBestBlue}\(\mathbf{100.0}_{\pm 0.0}\) & \(\underline{97.5}_{\pm 2.2}\) & \(\underline{93.1}_{\pm 0.7}\) & \(86.9_{\pm 0.6}\) & \(83.9_{\pm 2.2}\) & \cellcolor{ccpoBestBlue}\(\mathbf{75.0}_{\pm 0.8}\) & \cellcolor{ccpoBestBlue}\(\mathbf{85.4}_{\pm 1.6}\) \\
\ccpoSubgroupRule{11}
& \(\Delta\ \mathrm{GRPO}\)
    & \(16.6\text{--}21.6\) & \(12.1\text{--}18.2\) & \(8.7\text{--}33.5\) & \(20.3\text{--}22.8\) & \(19.3\) & \(8.9\text{--}22.0\) & \(18.2\text{--}31.3\) & \(12.3\text{--}18.5\) & \(13.1\text{--}14.8\) \\
\midrule
\multicolumn{11}{@{}l}{\textit{Qwen2.5-7B-Instruct}} \\
Prompting & Qwen2.5\textsuperscript{*}
    & \(19.3\) & \(37.1\) & \(38.5\) & \(14.8\) & \(12.5\) & \(8.0\) & \(4.2\) & \(3.0\) & \(26.6\) \\
Prompting & ReAct
    & \(31.2\) & \(48.5\) & \(35.4\) & \(34.3\) & \(13.2\) & \(18.2\) & \(17.6\) & \(19.5\) & \(46.2\) \\
Prompting & Reflexion
    & \(42.7\) & \(62.0\) & \(41.6\) & \(44.9\) & \(30.9\) & \(36.3\) & \(23.8\) & \(28.8\) & \(58.1\) \\
RL Training & PPO (with critic)
    & \(80.4_{\pm 2.7}\) & \(\underline{92.3}_{\pm 4.0}\) & \(64.0_{\pm 8.4}\) & \(92.5_{\pm 2.4}\) & \(89.5_{\pm 7.0}\) & \(80.3_{\pm 2.0}\) & \(68.8_{\pm 8.3}\) & \(68.7_{\pm 5.1}\) & \(81.4_{\pm 3.1}\) \\
RL Training & RLOO
    & \(75.5_{\pm 4.6}\) & \(87.6_{\pm 4.3}\) & \(78.2_{\pm 8.3}\) & \(87.3_{\pm 5.8}\) & \(81.3_{\pm 7.6}\) & \(71.9_{\pm 5.2}\) & \(48.9_{\pm 8.4}\) & \(65.7_{\pm 4.0}\) & \(80.3_{\pm 3.2}\) \\
RL Training & GRPO\textsuperscript{*}
    & \(82.0_{\pm 0.8}\) & \(85.8_{\pm 2.1}\) & \(71.9_{\pm 10.3}\) & \(92.1_{\pm 0.7}\) & \(84.2_{\pm 2.7}\) & \(81.2_{\pm 1.8}\) & \(69.1_{\pm 0.8}\) & \(72.9_{\pm 3.2}\) & \(82.9_{\pm 0.6}\) \\
RL Training & GiGPO\textsuperscript{*}
    & \(92.2_{\pm 0.8}\) & \cellcolor{ccpoBestBlue}\(\mathbf{100.0}_{\pm 0.0}\) & \(93.3_{\pm 11.5}\) & \(\underline{96.0}_{\pm 0.3}\) & \(\underline{95.7}_{\pm 3.7}\) & \(\underline{89.9}_{\pm 2.1}\) & \(76.5_{\pm 2.3}\) & \(74.7_{\pm 4.0}\) & \(84.3_{\pm 3.5}\) \\
\ccpoReported{RL Training} & \ccpoReported{GiGPO}
    & \ccpoReported{\(90.8_{\pm 1.3}\)} & \ccpoReported{\(97.7_{\pm 1.6}\)} & \ccpoReported{\(82.7_{\pm 7.9}\)} & \ccpoReported{\(98.8_{\pm 1.6}\)} & \ccpoReported{\(83.7_{\pm 7.2}\)} & \ccpoReported{\(89.3_{\pm 8.2}\)} & \ccpoReported{\(79.2_{\pm 6.6}\)} & \ccpoReported{\(72.8_{\pm 3.2}\)} & \ccpoReported{\(84.4_{\pm 2.9}\)} \\
RL Training & HGPO\textsuperscript{*}
    & \(\underline{95.8}_{\pm 0.5}\) & \cellcolor{ccpoBestBlue}\(\mathbf{100.0}_{\pm 0.0}\) & \(\underline{96.7}_{\pm 5.8}\) & \cellcolor{ccpoBestBlue}\(\mathbf{100.0}_{\pm 0.0}\) & \cellcolor{ccpoBestBlue}\(\mathbf{100.0}_{\pm 0.0}\) & \(85.6_{\pm 1.9}\) & \(\underline{92.6}_{\pm 2.6}\) & \(77.6_{\pm 4.3}\) & \(\underline{86.1}_{\pm 3.0}\) \\
\ccpoReported{RL Training} & \ccpoReported{HGPO}
    & \ccpoReported{\(95.4_{\pm 0.6}\)} & \multicolumn{6}{c|}{\ccpoReported{Unreported}} & \ccpoReported{\(78.5_{\pm 1.4}\)} & \ccpoReported{\(89.0_{\pm 1.0}\)} \\
RL Training & G2PO\textsuperscript{*}
    & \(94.0_{\pm 0.5}\) & \cellcolor{ccpoBestBlue}\(\mathbf{100.0}_{\pm 0.0}\) & \cellcolor{ccpoBestBlue}\(\mathbf{100.0}_{\pm 0.0}\) & \(\underline{96.0}_{\pm 0.3}\) & \(\underline{95.7}_{\pm 3.7}\) & \(85.6_{\pm 1.9}\) & \(88.2_{\pm 2.7}\) & \(72.4_{\pm 3.5}\) & \(83.8_{\pm 2.6}\) \\
\ccpoReported{RL Training} & \ccpoReported{G2PO}
    & \ccpoReported{\(96.9_{\pm 1.3}\)} & \ccpoReported{\(98.7_{\pm 2.3}\)} & \ccpoReported{\(90.6_{\pm 11.5}\)} & \ccpoReported{\(97.6_{\pm 2.1}\)} & \ccpoReported{\(100.0_{\pm 0.0}\)} & \ccpoReported{\(98.8_{\pm 2.1}\)} & \ccpoReported{\(93.7_{\pm 8.4}\)} & \ccpoReported{\(78.3_{\pm 0.6}\)} & \ccpoReported{\(89.8_{\pm 1.7}\)} \\
\ccpoSubgroupRule{11}
RL Training & \textbf{CCPO (ours)} & \multicolumn{9}{c}{} \\
& \(A_{\mathrm{CC}}\) only
    & \cellcolor{ccpoBestBlue}\(\mathbf{97.4}_{\pm 0.5}\) & \cellcolor{ccpoBestBlue}\(\mathbf{100.0}_{\pm 0.0}\) & \cellcolor{ccpoBestBlue}\(\mathbf{100.0}_{\pm 0.0}\) & \cellcolor{ccpoBestBlue}\(\mathbf{100.0}_{\pm 0.0}\) & \(\underline{95.7}_{\pm 3.7}\) & \cellcolor{ccpoBestBlue}\(\mathbf{94.3}_{\pm 2.2}\) & \cellcolor{ccpoBestBlue}\(\mathbf{94.1}_{\pm 2.5}\) & \cellcolor{ccpoBestBlue}\(\mathbf{79.4}_{\pm 1.6}\) & \cellcolor{ccpoBestBlue}\(\mathbf{89.4}_{\pm 1.0}\) \\
\addlinespace[2pt]
& \(A_{\mathrm{CC}} + A_{\mathrm{EP}}\)
    & \(94.3_{\pm 0.5}\) & \cellcolor{ccpoBestBlue}\(\mathbf{100.0}_{\pm 0.0}\) & \(85.9_{\pm 5.1}\) & \(\underline{96.0}_{\pm 0.3}\) & \cellcolor{ccpoBestBlue}\(\mathbf{100.0}_{\pm 0.0}\) & \(\underline{89.9}_{\pm 2.1}\) & \(88.2_{\pm 2.7}\) & \(\underline{78.4}_{\pm 3.9}\) & \(84.6_{\pm 3.9}\) \\
\ccpoSubgroupRule{11}
& \(\Delta\ \mathrm{GRPO}\)
    & \(12.3\text{--}15.4\) & \(14.2\) & \(14.0\text{--}28.1\) & \(3.9\text{--}7.9\) & \(11.5\text{--}15.8\) & \(8.7\text{--}13.1\) & \(19.1\text{--}25.0\) & \(5.5\text{--}6.5\) & \(1.7\text{--}6.5\) \\
\bottomrule
\end{tabular}%
}
\endgroup
\end{table}

\subsection{Ablation Studies}

We ablate CCPO with Qwen2.5-1.5B-Instruct, using $A_{\mathrm{CC}}$ on ALFWorld and $A_{\mathrm{CC}}+A_{\mathrm{EP}}$ on WebShop as reference configurations. All ablations follow the main evaluation protocol. 

\begin{table}[t]
\vspace{-4pt}
\centering
\begingroup
\definecolor{ccpoMetricGray}{HTML}{f0f0f0}
\definecolor{ccpoBestBlue}{HTML}{b9daf6}
\newcommand{\ccpoWindow}[2]{%
    \mbox{\begingroup
        \count255=#1\relax
        \loop\ifnum\count255>0
            \ccpoHistory\advance\count255 by -1\relax
        \repeat
        \ccpoAnchor
        \count255=#2\relax
        \loop\ifnum\count255>0
            \ccpoFuture\advance\count255 by -1\relax
        \repeat
    \endgroup}%
}
\setlength{\fboxsep}{1.5pt}
\caption{\textbf{Ablations with Qwen2.5-1.5B-Instruct on ALFWorld and WebShop.} \protect\colorbox{ccpoBestBlue}{\textbf{best}} and \underline{second best} are marked.
$H$ and $F$ denote active historical and future credit spans, with $W=H+F+1$.
Zero marks a disabled credit channel. Channel-only rows retain $K_H=K_F=2$.
Temporal-window rows also change the actual window lengths (Table~\ref{tab:ccpo-ablation-config}).
Markers: \protect\ccpoAnchor~current, \protect\ccpoHistory~history, and \protect\ccpoFuture~future.}
\label{tab:ablation-results-by-variant}
\setlength{\tabcolsep}{3pt}
\renewcommand{\arraystretch}{1.15}
\fontsize{8}{9}\selectfont
\ccpoMeasure{%
\begin{tabular}{@{}l>{\columncolor{ccpoMetricGray}}c!{\hspace{2pt}{\color{black!35}\vrule width 0.35pt}\hspace{2pt}}*{6}{c}|>{\columncolor{ccpoMetricGray}}c>{\columncolor{ccpoMetricGray}}c}
\toprule
\textbf{Method} & \multicolumn{7}{c|}{\textbf{ALFWorld}} & \multicolumn{2}{c}{\textbf{WebShop}} \\
& \textbf{All} & Pick & Look & Clean & Heat & Cool & Pick2 & Succ. & Score \\
\midrule
\textit{Qwen2.5-1.5B-Instruct} & \multicolumn{7}{c|}{\(A_{\mathrm{CC}}\) only} & \multicolumn{2}{c}{\(A_{\mathrm{CC}} + A_{\mathrm{EP}}\)} \\
w/o self-trajectory peer exclusion
    & \(90.6_{\pm 0.8}\) & \(97.0_{\pm 0.1}\) & \(85.9_{\pm 5.1}\) & \(88.1_{\pm 1.0}\) & \cellcolor{ccpoBestBlue}\(\mathbf{100.0}_{\pm 0.0}\) & \(86.9_{\pm 0.6}\) & \(83.8_{\pm 2.8}\) & \(56.0_{\pm 5.5}\) & \(75.1_{\pm 2.6}\) \\
w/o context summary
    & \(\underline{94.3}_{\pm 0.5}\) & \(97.0_{\pm 0.1}\) & \(81.9_{\pm 7.1}\) & \(92.1_{\pm 0.7}\) & \(\underline{95.7}_{\pm 3.7}\) & \(\underline{98.6}_{\pm 2.4}\) & \(\underline{92.6}_{\pm 2.6}\) & \(73.7_{\pm 3.5}\) & \(84.5_{\pm 3.0}\) \\
Uniform weighting
    & \(93.2_{\pm 0.5}\) & \(94.9_{\pm 1.9}\) & \cellcolor{ccpoBestBlue}\(\mathbf{92.6}_{\pm 6.4}\) & \(\underline{96.1}_{\pm 0.3}\) & \(\underline{95.7}_{\pm 3.7}\) & \(87.0_{\pm 0.6}\) & \(\underline{92.6}_{\pm 2.7}\) & \(72.4_{\pm 3.9}\) & \(83.1_{\pm 3.4}\) \\
Cosine-based context similarity
    & \(92.7_{\pm 0.5}\) & \(\underline{97.9}_{\pm 1.8}\) & \(\underline{89.3}_{\pm 0.6}\) & \(92.1_{\pm 0.7}\) & \(\underline{95.7}_{\pm 3.7}\) & \(89.9_{\pm 2.1}\) & \(88.2_{\pm 2.7}\) & \(70.3_{\pm 2.7}\) & \cellcolor{ccpoBestBlue}\(\mathbf{86.6}_{\pm 1.4}\) \\
\ccpoSubgroupRule{10}
\multicolumn{10}{@{}l}{\textbf{Historical credit only}} \\
H: 2, F: 0, W: 3\quad\ccpoWindow{2}{0}
    & \(91.4_{\pm 0.8}\) & \(93.9_{\pm 0.2}\) & \(85.9_{\pm 5.1}\) & \cellcolor{ccpoBestBlue}\(\mathbf{100.0}_{\pm 0.0}\) & \(93.1_{\pm 0.7}\) & \(86.9_{\pm 0.6}\) & \(83.8_{\pm 2.8}\) & \(72.7_{\pm 2.8}\) & \(85.1_{\pm 0.4}\) \\
\ccpoSubgroupRule{10}
\multicolumn{10}{@{}l}{\textbf{Future credit only}} \\
H: 0, F: 2, W: 3\quad\ccpoWindow{0}{2}
    & \(84.6_{\pm 1.6}\) & \(91.9_{\pm 1.9}\) & \(71.9_{\pm 10.3}\) & \(88.1_{\pm 1.0}\) & \(82.0_{\pm 2.4}\) & \(91.3_{\pm 0.4}\) & \(70.6_{\pm 3.0}\) & \(\underline{74.0}_{\pm 2.4}\) & \(85.1_{\pm 2.4}\) \\
\ccpoSubgroupRule{10}
\multicolumn{10}{@{}l}{\textbf{Symmetric context variation}} \\
H: 1, F: 1, W: 3\quad\ccpoWindow{1}{1}
    & \(86.7_{\pm 0.0}\) & \(91.9_{\pm 1.9}\) & \(78.5_{\pm 1.3}\) & \(92.1_{\pm 0.7}\) & \(91.0_{\pm 3.1}\) & \(85.6_{\pm 1.9}\) & \(75.0_{\pm 2.9}\) & \textemdash & \textemdash \\
H: 3, F: 3, W: 7\quad\ccpoWindow{3}{3}
    & \(91.4_{\pm 0.8}\) & \(97.0_{\pm 0.1}\) & \(85.9_{\pm 5.1}\) & \(96.0_{\pm 0.3}\) & \(93.1_{\pm 0.7}\) & \(91.3_{\pm 0.4}\) & \(79.4_{\pm 2.8}\) & \textemdash & \textemdash \\
\ccpoSubgroupRule{10}
\multicolumn{10}{@{}l}{\textbf{Asymmetric context variation}} \\
H: 4, F: 0, W: 5\quad\ccpoWindow{4}{0}
    & \(\underline{94.3}_{\pm 0.9}\) & \(97.0_{\pm 0.1}\) & \(\underline{89.3}_{\pm 0.6}\) & \cellcolor{ccpoBestBlue}\(\mathbf{100.0}_{\pm 0.0}\) & \(\underline{95.7}_{\pm 3.7}\) & \(91.3_{\pm 0.4}\) & \(88.2_{\pm 2.7}\) & \textemdash & \textemdash \\
H: 3, F: 1, W: 5\quad\ccpoWindow{3}{1}
    & \(92.7_{\pm 0.9}\) & \cellcolor{ccpoBestBlue}\(\mathbf{100.0}_{\pm 0.0}\) & \(\underline{89.3}_{\pm 0.6}\) & \(96.0_{\pm 0.3}\) & \(91.0_{\pm 3.1}\) & \cellcolor{ccpoBestBlue}\(\mathbf{100.0}_{\pm 0.0}\) & \(73.6_{\pm 3.9}\) & \textemdash & \textemdash \\
H: 1, F: 3, W: 5\quad\ccpoWindow{1}{3}
    & \(89.3_{\pm 1.6}\) & \(93.9_{\pm 0.2}\) & \(82.6_{\pm 10.9}\) & \(89.6_{\pm 1.5}\) & \(82.0_{\pm 2.4}\) & \(91.3_{\pm 0.4}\) & \(88.2_{\pm 2.7}\) & \textemdash & \textemdash \\
H: 0, F: 4, W: 5\quad\ccpoWindow{0}{4}
    & \(62.0_{\pm 1.8}\) & \(71.7_{\pm 2.6}\) & \(82.6_{\pm 10.9}\) & \(58.2_{\pm 6.1}\) & \(66.1_{\pm 3.4}\) & \(59.4_{\pm 1.3}\) & \(44.1_{\pm 1.1}\) & \textemdash & \textemdash \\
\ccpoSubgroupRule{10}
\multicolumn{10}{@{}l}{\textbf{CCPO (ours)}} \\
H: 2, F: 2, W: 5\quad\ccpoWindow{2}{2}
    & \cellcolor{ccpoBestBlue}\(\mathbf{96.9}_{\pm 0.8}\) & \cellcolor{ccpoBestBlue}\(\mathbf{100.0}_{\pm 0.0}\) & \(75.2_{\pm 4.5}\) & \cellcolor{ccpoBestBlue}\(\mathbf{100.0}_{\pm 0.0}\) & \(93.1_{\pm 0.7}\) & \cellcolor{ccpoBestBlue}\(\mathbf{100.0}_{\pm 0.0}\) & \cellcolor{ccpoBestBlue}\(\mathbf{97.0}_{\pm 2.6}\) & \cellcolor{ccpoBestBlue}\(\mathbf{75.0}_{\pm 0.8}\) & \(\underline{85.4}_{\pm 1.6}\) \\
\bottomrule
\end{tabular}%
}
\ifdim\wd0>\linewidth
    \resizebox{\linewidth}{!}{\usebox{0}}%
\else
    \usebox{0}%
\fi
\endgroup
\end{table}

\vspace{-1pt}
\paragraph{Peer selection and contextual weighting.}
Relative to CCPO, including same-trajectory peers lowers mean success by 6.3 on ALFWorld and 19.0 percentage points on WebShop (Table~\ref{tab:ablation-results-by-variant}). Uniform weighting lowers success by 3.7 and 2.6 points, and cosine weighting by 4.2 and 4.7. 
\vspace{-1pt}
\paragraph{Structured context summaries alter peer influence.}
Context summaries add accumulated trajectory statistics to semantic representations: step count, distinct-observation count, novelty fraction, and a revisit indicator. Removing them lowers mean success by 2.6 percentage points on ALFWorld and 1.3 on WebShop (Table~\ref{tab:ablation-results-by-variant}). Adding summaries changes normalized weights by more than one percentage point for 33.8\% of matched query--peer pairs (Table~\ref{tab:peer-reweighting}; Appendix~\ref{app:stats:table3}).

\vspace{-4pt}
\begin{table}[ht]
\vspace{-4pt}
\caption{\textbf{Absolute changes in matched peer weights from context summaries.}}
\label{tab:peer-reweighting}
\begin{center}
\scriptsize
\setlength{\tabcolsep}{3pt}
\begin{tabular}{lccccc}
\toprule
Change (pp) & $>0.25$ & $>0.5$ & $>1$ & $>5$ & $>10$ \\
\midrule
Pairs (\%) & 62.1 & 48.7 & 33.8 & 8.6 & 3.6 \\
\bottomrule
\end{tabular}
\end{center}
\end{table}

\vspace{-1pt}
\paragraph{Historical and future credit.}
We compare the combined, history-only, and future-only credit configurations, retaining episode-level credit on WebShop. These channel-only ablations switch off the other credit component while retaining the default prompt-history length and future horizon, $K_{\mathrm H}=K_{\mathrm F}=2$ (Table~\ref{tab:ccpo-ablation-config}). On ALFWorld, success falls from 96.9\% to 91.4\% when using only historical credit and 84.6\% when using only future credit. On WebShop, it decreases from 75.0\% to 72.7\% and 74.0\%, respectively (Table~\ref{tab:ablation-results-by-variant}). The combined configuration achieves the highest mean success on both benchmarks. History-only exceeds future-only in mean success on ALFWorld but trails it on WebShop, suggesting task-dependent credit utility.

\vspace{-1pt}
\paragraph{Temporal window allocation.}
We vary symmetric windows on history length and future horizon. Increasing the symmetric window from $(2,2)$ to $(3,3)$ reduces success from 96.9\% to 91.4\%. We then compare asymmetric temporal allocations at fixed context span. The default CCPO (2 steps each way) outperforms all tested asymmetric windows. We restrict these variations to ALFWorld, where successful 1.5B trajectories average 8.7 steps versus 5.1 on WebShop under the reference configurations (Appendix~\ref{app:trajectory-lengths}, Figure~\ref{fig:trajectory-lengths}). These results favor a compact yet rich context window that balances locality with context span, since two steps each way outperforms both one and three steps.

\subsection{Contextual Credit Analysis}

\begin{figure}[!ht]
    \centering
    \vspace{-4pt}
    \includegraphics[width=\linewidth]{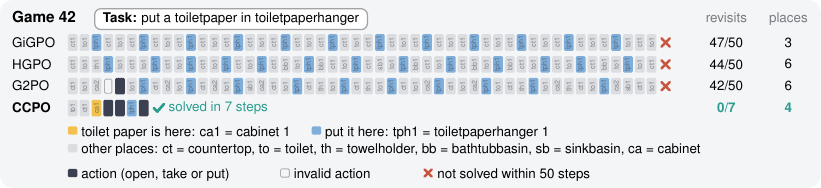}
    \caption{\textbf{ALFWorld evaluation trajectories (Qwen2.5-1.5B-Instruct).} Each cell denotes one step. Revisit counts track navigation to previously visited locations. CCPO uses $A_{\mathrm{CC}}$ only.}
    \label{fig:revisit-ex}

    \vspace{4pt}
    \captionof{table}{\textbf{Revisit rates on ALFWorld evaluation rollouts.}}
\label{tab:eval-revisit-rate}
\begin{center}
\scriptsize
\setlength{\tabcolsep}{3pt}
\begin{tabular}{lcccc}
\toprule
Method & GiGPO & G2PO & HGPO & CCPO ($A_{\mathrm{CC}}$ only) \\
\midrule
Revisit rate (\%) & $41.0_{\pm 0.9}$ & $26.8_{\pm 0.7}$ & $36.4_{\pm 0.7}$ & $\mathbf{22.5}_{\pm 1.8}$ \\
\bottomrule
\end{tabular}
\end{center}

    \vspace{4pt}
\end{figure}

\paragraph{Identical observations can conceal different task progress.}
\looseness=-1
In Figure~\ref{fig:revisit-ex}, GiGPO and G2PO repeatedly visit an empty toiletpaper hanger without acquiring the target. CCPO encounters the same observation after acquisition and completes the task (Appendix~\ref{app:failure-revisits}). Appendix~\ref{app:failure-subgoal} presents a complementary failure mode: G2PO repeatedly removes and replaces one spraybottle, whereas CCPO places two distinct objects (Figure~\ref{fig:revisit-subgoal}). Across three evaluation draws, CCPO's revisit rate is 22.5\%, versus 26.8--41.0\% for the baselines (Table~\ref{tab:eval-revisit-rate}).

\begin{figure}[!ht]
    \centering
    {\footnotesize\bfseries
    \makebox[0.5\linewidth][c]{(a) Contextual vs. episode advantage}%
    \makebox[0.5\linewidth][c]{(b) Credit decomposition}\par}
    \smallskip
    \vspace{-4pt}
    \includegraphics[width=\linewidth]{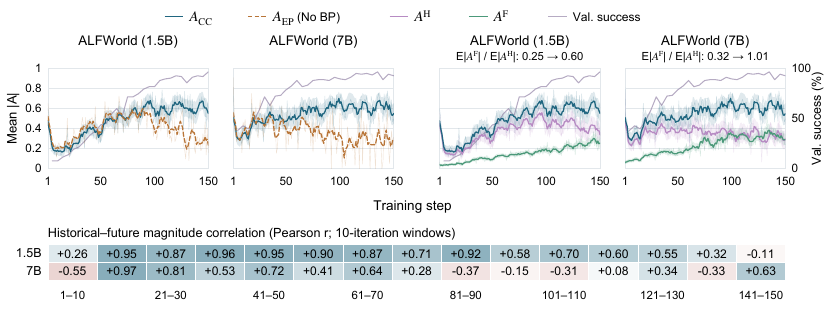}
    \caption{\textbf{Contextual credit dynamics on ALFWorld (Qwen2.5-1.5B and 7B).}
    (a) $A_{\mathrm{CC}}$ versus $A_{\mathrm{EP}}$ from the same rollouts. Only $A_{\mathrm{CC}}$ is backpropagated.
    (b) Historical, future, and combined credit. Future-to-history ratios compare iterations 1--30 and 121--150.
    The heatmap correlates unsmoothed historical and future magnitudes by non-overlapping 10-iteration windows.
    Curves: mean absolute credit, centered 5-iteration means, and faint raw traces. Bands: 95\% trajectory-bootstrap CIs.}
    \label{fig:ccpo-credit-dynamics}
    \label{fig:ccpo-acc-vs-aep}
    \label{fig:ccpo-credit-decomp}
    \vspace{-5pt}
\end{figure}

\vspace{-1pt}
\paragraph{Contextual credit provides unique learning signals.}
Figure~\ref{fig:ccpo-credit-dynamics}a compares $A_{\mathrm{CC}}$ with counterfactual $A_{\mathrm{EP}}$ from the same $A_{\mathrm{CC}}$-only ALFWorld rollouts. $A_{\mathrm{EP}}$ is logged for comparison and excluded from policy updates. Their mean absolute magnitudes diverge as validation success improves: $A_{\mathrm{EP}}$ diminishes while $A_{\mathrm{CC}}$ remains elevated, so a learning signal remains after most episodes succeed.

\vspace{-1pt}
\paragraph{The composition of contextual advantage changes during training.}
Figure~\ref{fig:ccpo-credit-dynamics}b shows historical credit dominating initially, with future credit becoming more prominent despite fixed combination weights. Between iterations 1--30 and 121--150, the future-to-history mean absolute magnitude ratio rises from 0.25 to 0.60 at 1.5B and from 0.32 to 1.01 at 7B. Correlations between historical and future magnitudes within 10-iteration windows become less consistently positive later, especially at 7B. The two components thus evolve differently across training, supporting complementary roles.

\begin{figure}[!ht]
    \centering
    \vspace{-4pt}
    \includegraphics[width=0.70\linewidth]{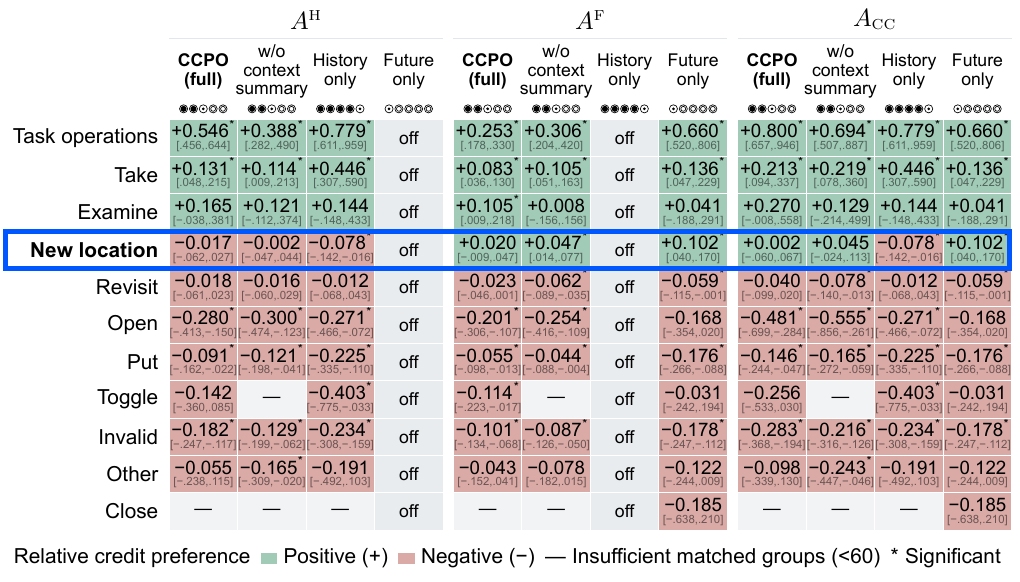}
    \caption{\textbf{Credit preferences on ALFWorld (Qwen2.5-1.5B-Instruct).} Mean category-relative credit within matched task--observation groups and iterations, from successful rollouts at iterations 101--150. Asterisks: Bonferroni-adjusted CIs excluding zero. Dashes: fewer than 60 groups. \texttt{Off}: inactive component. Context markers denote \ccpoAnchor~current, \ccpoHistory~history, and \ccpoFuture~future.}
    \label{fig:ccpo-action-analysis}
    \vspace{-4pt}
\end{figure}

\vspace{-1pt}
\paragraph{What does future credit add beyond historical credit?}
\looseness=-1
Using successful 1.5B ALFWorld rollouts from iterations 101--150, we compare category-relative credit within task--observation-matched groups per iteration, with bootstrap CIs and Bonferroni correction across ten categories (Figure~\ref{fig:ccpo-action-analysis}). Filtering and statistical procedures are detailed in Appendix~\ref{app:stats:fig4a}. Relative to other actions, $A^{\mathrm F}$ favors new-location transitions more than $A^{\mathrm H}$ (difference: $+0.037$; adjusted CI: $[0.001, 0.073]$), while both favor cleaning, heating, and cooling. The $(H,F)=(4,0)$, $(0,4)$, and w/o-context-summary variants show similar patterns. Paired action timing also differs: against full CCPO, the future-only policy first opens a receptacle earlier in 72.1\% of paired games, and the history-only policy later in 54.3\% (Appendix~\ref{app:stats:fig4b}, Figure~\ref{fig:ccpo-action-timing}). The failure case in Figure~\ref{fig:revisit-ex} shows that identical observations can accompany different task progress; the credit preferences in Figure~\ref{fig:ccpo-action-analysis} complement this behavioral evidence by showing how historical and future credit evaluate exploratory transitions differently.

\section{Conclusion}

We introduced Contextual Credit Policy Optimization (CCPO) for two challenges in comparison-based step-level credit: whom a step is compared with, and what it is credited for. CCPO weights observation-matched peers by contextual similarity and pairs historical credit with future credit, both computed from collected rollouts without an auxiliary value critic. With Qwen2.5-7B-Instruct, it reaches 97.4\% success on ALFWorld and 79.4\% on WebShop. In ablations, uniform weights lower success by 3.7 and 2.6 points, and either credit alone lowers it by up to 12.3.

\vspace{-1pt}
\paragraph{Limitations.} CCPO has three main limitations. First, our evaluation covers two multi-step benchmarks and one model family. Broader generalization and applicability to short-horizon or single-step tasks remain untested. Second, exact observation matching can limit peer availability, while fixed temporal windows and hand-designed trajectory statistics may require adaptation across environments. Finally, future credit vanishes when all rollouts in a task group receive zero terminal reward, limiting the future channel's contribution under sparse rewards, particularly early in training when successful trajectories are scarce.

\vspace{-1pt}
\paragraph{Future work.}
The changing balance between historical and future credit motivates adaptive horizons and stage-dependent weighting. Further directions include streaming credit estimation from incomplete rollouts, multimodal contexts combining visual observations and textual histories, semantic peer retrieval beyond exact observation matching, and uncertainty estimates for correlated visits. Broader evaluation should examine generalization across model families and domains, computational scaling, and the effects of realized-future conditioning on policy updates.

\clearpage
\bibliography{references}
\bibliographystyle{assets/plainnat}

\clearpage
\beginappendix
\setcounter{table}{0}
\renewcommand{\thetable}{S\arabic{table}}
\renewcommand{\theHtable}{supp.\arabic{table}}
\setcounter{figure}{0}
\renewcommand{\thefigure}{S\arabic{figure}}
\renewcommand{\theHfigure}{supp.\arabic{figure}}

\section{Notation}
\label{s01-notation}

\begingroup
\small
\setlength{\tabcolsep}{5pt}
\setcounter{LTchunksize}{200}
\renewcommand{\arraystretch}{1}
\setlength{\LTcapwidth}{\linewidth}
\begin{longtable}{@{}>{\raggedright\arraybackslash}p{0.30\linewidth}>{\raggedright\arraybackslash}p{\dimexpr0.70\linewidth-10pt\relax}@{}}
\caption{\textbf{Notation for CCPO.} Subscripts index trajectories and steps; $\mathrm H$ and $\mathrm F$ denote historical and future channels. Implementation details appear in Appendix~\ref{app:ccpo-implementation}.}\label{tab:ccpo-notation}\\
\toprule
\textbf{Symbol} & \textbf{Definition}\\
\midrule
\endfirsthead
\multicolumn{2}{@{}l}{\textit{Table \thetable\ continued}}\\
\toprule
\textbf{Symbol} & \textbf{Definition}\\
\midrule
\endhead
\midrule
\multicolumn{2}{r@{}}{\textit{Continued on next page}}\\
\endfoot
\bottomrule
\endlastfoot
\ccpoShadedRow\multicolumn{2}{@{}l@{}}{\strut\textbf{Tasks, trajectories, and temporal indexing}}\\*
\addlinespace[3pt]
$x,\ x_i$ & Task instance; $x_i$ is the task associated with trajectory $i$.\\
\addlinespace[1pt]
$N$ & Number of trajectories sampled per task in a rollout batch.\\
\addlinespace[1pt]
$\tau_i$ & Trajectory $i$, containing observations $o_{i,0:T_i}$, actions $a_{i,0:T_i-1}$, and their outcomes.\\
\addlinespace[1pt]
$i,\ j$ & Query-trajectory and peer-trajectory indices.\\
\addlinespace[1pt]
$t,\ s$ & Zero-based query-step and peer-step indices: $0\le t<T_i$, $0\le s<T_j$.\\
\addlinespace[1pt]
$T_i$ & Number of actions in trajectory $i$; $T_i$ also indexes its terminal endpoint.\\
\addlinespace[1pt]
$o_{i,t}$ & Observation before action $a_{i,t}$; matching uses exact string equality.\\
\addlinespace[1pt]
$a_{i,t}$ & Generated response, including reasoning and the executable action.\\
\addlinespace[1pt]
$p_{i,t}$ & Task, current observation, and up to $K_{\mathrm H}$ preceding observation--action pairs; excludes the current response.\\
\addlinespace[1pt]
$r^{\mathrm{env}}_{i,t},\ R_i$ & Environment reward after action $a_{i,t}$, and terminal trajectory outcome.\\
\addlinespace[1pt]
$\gamma,\ \eta$ & Return discount factor and current invalid-action penalty coefficient.\\
\addlinespace[1pt]
$K_{\mathrm H},\ K_{\mathrm F}$ & Prompt-history length and realized future horizon, both $2$ by default.\\
\addlinespace[1pt]
$t^+$ & Future endpoint on the collected trajectory: $t^+=\min(t+K_{\mathrm F},T_i)$.\\
\addlinespace[1pt]
$H,\ F,\ W$ & Active credit spans in the ablation table, with $W=H+F+1$; zero denotes a disabled channel. Channel-only rows retain $K_{\mathrm H}=K_{\mathrm F}=2$, while temporal-window variations use $K_{\mathrm H}=H$ and $K_{\mathrm F}=F$.\\
\ccpoShadedRow\multicolumn{2}{@{}l@{}}{\strut\textbf{Context representation}}\\*
\addlinespace[3pt]
$h_{i,t}$ & Frozen reference model\textquotesingle s final-layer hidden state at the last prompt token.\\
\addlinespace[1pt]
$\bar h,\ P_{\perp,3}$ & Batch mean hidden state and projection removing its three leading principal directions.\\
\addlinespace[1pt]
$n_{i,t}$ & Distinct-observation count: $|\{o_{i,0},\ldots,o_{i,t}\}|$.\\
\addlinespace[1pt]
$n_{i,t}/(t+1)$ & Observation-novelty ratio over the observed prefix.\\
\addlinespace[1pt]
$\mathbf 1\{\mathrm{revisit}_{i,t}\}$ & Whether the current observation appeared earlier in the same trajectory.\\
\addlinespace[1pt]
$c_{i,t}$ & 37-dimensional encoding of elapsed steps, distinct observations, novelty ratio, and revisit indicator.\\
\addlinespace[1pt]
$\nu(v)$ & L2 normalization: $v/\max(\|v\|_2,10^{-9})$.\\
\addlinespace[1pt]
$\phi_{i,t}$ & Fused context feature; see Equation~\ref{eq:context-representation}.\\
\addlinespace[1pt]
$[u;v]$ & Vector concatenation.\\
\ccpoShadedRow\multicolumn{2}{@{}l@{}}{\strut\textbf{Peer matching and contextual weighting}}\\*
\addlinespace[3pt]
$\mathcal E_{i,t}$ & Exact task--observation peers outside the query trajectory; Equation~\ref{eq:exact-peer-set}.\\
\addlinespace[1pt]
$\mathcal P_{i,t}$ & Peers used for readout: $\mathcal E_{i,t}$ when nonempty, otherwise all same-task nonterminal steps with $j\ne i$.\\
\addlinespace[1pt]
$u=(i,t),\ v=(j,s)$ & Shorthand for query and peer step identifiers.\\
\addlinespace[1pt]
$w_{u,v}$ & Kernel weight: $\exp(-\|\phi_u-\phi_v\|_2/\tau_B)$.\\
\addlinespace[1pt]
$\bar w_{u,v}$ & Query-normalized weight: $w_{u,v}/\sum_{v^{\prime}\in\mathcal P_{i,t}}w_{u,v^{\prime}}$.\\
\addlinespace[1pt]
$B$ & Observation or task bucket used to determine the bandwidth.\\
\addlinespace[1pt]
$m_B,\ \widetilde m_B$ & Median pairwise distance in $B$ and its safeguarded value; Appendix~\ref{app:ccpo-readouts}.\\
\addlinespace[1pt]
$\tau_B$ & Kernel bandwidth $0.15\widetilde m_B$, distinct from trajectory $\tau_i$.\\
\addlinespace[1pt]
$r_{j,s}$ & Readout target: $Y_{j,s}$ for return baselines or $Z_{j,s}$ for outcome values.\\
\addlinespace[1pt]
$\mathcal B_{i,t}[r]$ & Contextual weighted readout; Equation~\ref{eq:contextual-estimator}.\\
\ccpoShadedRow\multicolumn{2}{@{}l@{}}{\strut\textbf{Temporal credit signals}}\\*
\addlinespace[3pt]
$Y_{i,t}$ & Historical return target including the current invalid-action penalty; Equation~\ref{eq:historical-target}.\\
\addlinespace[1pt]
$\mathcal B_{i,t}[Y]$ & Contextual peer baseline for the realized return.\\
\addlinespace[1pt]
$C^{\mathrm H}_{i,t}$ & Historical credit: $Y_{i,t}-\mathcal B_{i,t}[Y]$.\\
\addlinespace[1pt]
$Z_{i,s}$ & Unpenalized outcome target: $\gamma^{T_i-s}R_i$.\\
\addlinespace[1pt]
$V_{i,s}$ & Contextual outcome value: $\mathcal B_{i,s}[Z]$ for $s<T_i$, and $R_i$ at $s=T_i$.\\
\addlinespace[1pt]
$\Delta V_{i,t}$ & Outcome-value change: $V_{i,t^+}-V_{i,t}$.\\
\addlinespace[1pt]
$C^{\mathrm F}_{i,t}$ & Task-standardized future credit: $z_{x_i}[\Delta V]_{i,t}$.\\
\ccpoShadedRow\multicolumn{2}{@{}l@{}}{\strut\textbf{Support, normalization, and advantages}}\\*
\addlinespace[3pt]
$\mathcal S_x^{\mathrm H}$ & Unique steps of task $x$ with a defined historical baseline.\\
\addlinespace[1pt]
$\mathcal S_x^{\mathrm F}$ & Unique task-$x$ steps with both outcome-value endpoints defined.\\
\addlinespace[1pt]
$\mathcal S_x,\ \widehat{\mathcal S}_x$ & $\mathcal S_x=\mathcal S_x^{\mathrm H}\cup\mathcal S_x^{\mathrm F}$; $\widehat{\mathcal S}_x$ is the corresponding multiset of restored training rows.\\
\addlinespace[1pt]
$\mathcal R_x$ & All restored input step rows of task $x$, used for episode statistics.\\
\addlinespace[1pt]
$\bar u_S,\ s_S(u)$ & Mean and sample standard deviation over set or row multiset $S$.\\
\addlinespace[1pt]
$z_S[u]$ & Standardization: $(u-\bar u_S)/(s_S(u)+10^{-6})$ for $|S|>1$.\\
\addlinespace[1pt]
$z_x[\Delta V]$ & Future-credit standardization over $\mathcal S_x^{\mathrm F}$.\\
\addlinespace[1pt]
$\mathcal N_x[\cdot]$ & Final contextual transformation.\\
\addlinespace[1pt]
$A_{\mathrm{CC},i,t}$ & Contextual advantage $\mathcal N_{x_i}[C^{\mathrm H}+C^{\mathrm F}]_{i,t}$, also denoted $A^{\mathrm{CC}}_{i,t}$.\\
\addlinespace[1pt]
$A^{\mathrm H}_{i,t}$ & Historical contribution after the final transformation; Equation~\ref{eq:app-ccpo-component-decomposition}.\\
\addlinespace[1pt]
$A^{\mathrm F}_{i,t}$ & Future contribution after the same transformation; $A_{\mathrm{CC}}=A^{\mathrm H}+A^{\mathrm F}$.\\
\addlinespace[1pt]
$\mu_x^d,\ \sigma_x$ & Component mean and common scale for the additive decomposition; Appendix~\ref{app:ccpo-normalization}.\\
\addlinespace[1pt]
$e_{i,t}$ & Episode-channel score: $R_i-\eta\mathbf1\{a_{i,t}\text{ invalid}\}$.\\
\addlinespace[1pt]
$A_{\mathrm{EP},i,t}$ & Episode-score advantage over $\mathcal R_x$.\\
\addlinespace[1pt]
$\alpha_{\mathrm{EP}}$ & Episode-channel weight: $0$ by default, $1$ for the augmented variant.\\
\addlinespace[1pt]
$A_{i,t}$ & Optimizer advantage: $A_{\mathrm{CC},i,t}+\alpha_{\mathrm{EP}}A_{\mathrm{EP},i,t}$.\\
\ccpoShadedRow\multicolumn{2}{@{}l@{}}{\strut\textbf{Policy optimization}}\\*
\addlinespace[3pt]
$\theta,\ \pi_\theta$ & Trainable policy parameters and the current LLM policy.\\
\addlinespace[1pt]
$\pi_{\mathrm{old}}$ & Behavior policy used to collect rollouts.\\
\addlinespace[1pt]
$\pi_{\mathrm{ref}}$ & Frozen reference policy for feature extraction and KL regularization.\\
\addlinespace[1pt]
\end{longtable}
\endgroup

\section{Training and Ablation Configurations}
\label{app:training-config}
\begingroup
\small
\setlength{\tabcolsep}{2.5pt}
\setlength{\LTcapwidth}{\linewidth}
\begin{longtable}{@{}>{\raggedright\arraybackslash}p{0.31\linewidth}*{5}{>{\centering\arraybackslash}p{\dimexpr0.138\linewidth-2\tabcolsep\relax}}@{}}
\caption{\textbf{Training configurations shared by 1.5B and 7B models on ALFWorld and WebShop.}}
\label{tab:training-config}\\
\toprule
Setting & GRPO & GiGPO & HGPO & G2PO & CCPO \\
\midrule
\endfirsthead
\multicolumn{6}{@{}l}{\textit{Table \thetable\ continued}}\\
\toprule
Setting & GRPO & GiGPO & HGPO & G2PO & CCPO \\
\midrule
\endhead
\bottomrule
\endfoot
\bottomrule
\endlastfoot
\ccpoShadedRow\multicolumn{6}{@{}l@{}}{\strut\textbf{Both benchmarks}} \\*
Training iterations & 150 & 150 & 150 & 150 & 150 \\
Tasks $\times$ rollouts per iteration & $16\times8$ & $16\times8$ & $16\times8$ & $16\times8$ & $16\times8$ \\
Total sampled trajectories & 19,200 & 19,200 & 19,200 & 19,200 & 19,200 \\
NVIDIA H200 GPUs per run & 8 & 8 & 8 & 8 & 8 \\
Learning rate & $10^{-6}$ & $10^{-6}$ & $10^{-6}$ & $10^{-6}$ & $10^{-6}$ \\
Discount factor & 0.95 & 0.95 & 0.95 & 0.95 & 0.95 \\
Actor KL coefficient & 0.01 & 0.01 & 0.01 & 0.01 & 0.01 \\
KL estimator & \texttt{low\_var\_kl} & \texttt{low\_var\_kl} & \texttt{low\_var\_kl} & \texttt{low\_var\_kl} & \texttt{low\_var\_kl} \\
KL included in reward & No & No & No & No & No \\
Invalid-action penalty coeff. & 0.1 & 0.1 & 0.1 & 0.1 & 0.1 \\
Environment training seed & 0 & 0 & 0 & 0 & 0 \\
Prompt history (steps) & 2 & 2 & 2 & 2 & 2 \\
Response token limit / step & 512 & 512 & 512 & 512 & 512 \\
Prompt truncation & \texttt{error} & \texttt{error} & \texttt{left} & \texttt{error} & \texttt{error} \\
Validation decoding & Sampling & Sampling & Sampling & Sampling & Sampling \\
Validation temperature & 0.4 & 0.4 & 0.4 & 0.4 & 0.4 \\
Validation interval (iterations) & 5 & 5 & 5 & 5 & 5 \\
\midrule
\ccpoShadedRow\multicolumn{6}{@{}l@{}}{\strut\textbf{ALFWorld}} \\*
PPO mini-batch size & 256 & 256 & 256 & 256 & 256 \\
Maximum interaction steps & 50 & 50 & 50 & 50 & 50 \\
Prompt token limit & 2,048 & 2,048 & 4,096 & 2,048 & 2,048 \\
\midrule
\ccpoShadedRow\multicolumn{6}{@{}l@{}}{\strut\textbf{WebShop}} \\*
PPO mini-batch size & 64 & 64 & 64 & 64 & 64 \\
Maximum interaction steps & 15 & 15 & 30 & 15 & 15 \\
Prompt token limit & 4,096 & 4,096 & 4,096 & 4,096 & 4,096 \\
\end{longtable}
\endgroup

\begingroup
\small
\setlength{\tabcolsep}{4pt}
\setlength{\LTcapwidth}{\linewidth}
\begin{longtable}{@{}p{0.36\linewidth}p{\dimexpr0.64\linewidth-2\tabcolsep\relax}@{}}
\caption{\textbf{CCPO variant and ablation configurations.}
Each ablation changes only the listed settings relative to its
corresponding reference configuration. The default uses prompt history
$K_{\mathrm H}=2$, future horizon $K_{\mathrm F}=2$, and unit weights for both
credit channels. In Table~\ref{tab:ablation-results-by-variant}, a disabled
channel has active span zero; its underlying window is retained in the
channel-only ablations.}
\label{tab:ccpo-ablation-config}\\
\toprule
Variant & Configuration change \\
\midrule
\endfirsthead
\multicolumn{2}{@{}l}{\textit{Table \thetable\ continued}}\\
\toprule
Variant & Configuration change \\
\midrule
\endhead
\bottomrule
\endfoot
\bottomrule
\endlastfoot
$A_{\mathrm{CC}}+A_{\mathrm{EP}}$
    & Add episode advantage with unit weight after contextual
      normalization, without renormalizing the sum. \\[3pt]

w/o self-trajectory peer exclusion
    & Allow matching steps from the query trajectory,
      including the query itself, as peers. \\[3pt]

w/o context summary
    & Set the accumulated-statistics weight from 1 to 0;
      retain frozen hidden-state features and prompt history. \\[3pt]

Cosine-based context similarity
    & Replace the exponential kernel with cosine-based weighting. \\[3pt]

Historical credit only
    & Turn off future credit by setting its weight from 1 to 0.
      Retain historical credit with unit weight and the default windows
      $K_{\mathrm H}=K_{\mathrm F}=2$; the active spans are $(H,F)=(2,0)$. \\[3pt]

Future credit only
    & Turn off historical credit by setting its weight from 1 to 0.
      Retain future credit with unit weight and horizon $K_{\mathrm F}=2$,
      together with prompt history $K_{\mathrm H}=2$;
      the active spans are $(H,F)=(0,2)$. \\[3pt]

Temporal window variation
    & Change the actual prompt-history length and future horizon to the
      listed values: $K_{\mathrm H}=H$ and $K_{\mathrm F}=F$.
      Set a channel's weight to 0 when its listed span is zero, and to 1
      otherwise. For example, $(H,F)=(4,0)$ uses four history steps and
      turns off future credit; $(0,4)$ uses a four-step future horizon,
      no preceding prompt-history pairs, and turns off historical credit. \\
\end{longtable}
\endgroup

\section{CCPO Implementation Details}
\label{app:ccpo-implementation}

Using complete, globally gathered rollouts, we remove identical step copies
before feature construction, peer readouts, and future-credit standardization.
Original rows are restored for final advantage computation before PPO
minibatching. Training and ablation configurations are in
Appendix~\ref{app:training-config}.

\subsection{Context Representation}
\label{app:ccpo-features}
The frozen reference model supplies the final-layer hidden state at the last
prompt token. Across unique steps in the batch, we center these states, remove
their three leading principal directions, and L2-normalize. The summary encodes
elapsed steps $t$, distinct-observation count $n_{i,t}$, and novelty ratio
$n_{i,t}/(t+1)$ in three 12-dimensional thermometer blocks with upper bounds
$30$, $25$, and $1$. For value $v$ and bound $b$, the first
$\lfloor12\operatorname{clip}(v/b,0,1)\rfloor$ coordinates are one; the rest are zero.
Appending the binary revisit indicator gives $c_{i,t}\in\mathbb R^{37}$.
With $\nu(v)=v/\max(\|v\|_2,10^{-9})$, the fused feature is
\begin{equation}
\phi_{i,t}=\nu\!\left([\nu(P_{\perp,3}(h_{i,t}-\bar h));\nu(c_{i,t})]\right).
\label{eq:context-representation}
\end{equation}
The projection is recomputed per batch without reward labels. With two or
three unique steps, we skip principal-direction removal; with one, we use the
raw hidden block. Nonterminal endpoints use their own prompt features
and accumulated statistics.

\subsection{Peer Readouts}
\label{app:ccpo-readouts}
Readouts use $\mathcal P_{i,t}$ from Section~\ref{sec:contextual-peer-estimation};
a step whose peer set is empty receives zero contextual credit. For the full observation bucket $B$, let $m_B$
be the median distance over distinct unordered pairs before trajectory
exclusion. Equation~\ref{eq:contextual-kernel} uses
$\tau_B=0.15\widetilde m_B$, with $\widetilde m_B=m_B$ when $m_B>10^{-9}$
and $1$ otherwise. Task-level fallback recomputes this bandwidth in the full task bucket.

Repeated visits retain individual weights. Each peer trajectory contributes
its weighted mean, weighted again by its total kernel mass. Both denominators
are floored at $10^{-12}$; all-zero weights yield zero while retaining support.
Away from these floors, this implements Equation~\ref{eq:contextual-estimator}.

\subsection{Credit Construction and Normalization}
\label{app:ccpo-normalization}
With terminal outcomes $R_i\in\{0,10\}$, the supplied historical target is
$Y_{i,t}=\gamma^{T_i-1-t}R_i-\eta\mathbf1\{a_{i,t}\text{ invalid}\}$,
whereas future values use $Z_{i,t}=\gamma^{T_i-t}R_i$ without penalties.
The current-step penalty enters $Y$ once.
At $t^+=\min(t+K_{\mathrm F},T_i)$, terminal values equal $R_i$, including
zero for unsuccessful rollouts. Both value endpoints must be defined; their
difference has no additional discount factor.

Let $h=t^+-t$. Equation~\ref{eq:future-credit} can be written as
\begin{equation}
\Delta V_{i,t}
=\bigl(\gamma^h V_{i,t^+}-V_{i,t}\bigr)
 +(1-\gamma^h)V_{i,t^+}.
\label{eq:app-future-discount-decomposition}
\end{equation}
Our value targets are nonnegative, so the second term is nonnegative
and can vary across steps. This time-dependent contribution can persist after task-wise centering.

Standardization uses sample standard deviation plus $10^{-6}$, with equal
weight per included row. Future differences are standardized over supported
unique steps within each task in both benchmarks; fewer than two eligible
steps give zero future credit. Unavailable components are zero-filled before
combination. Unsupported contextual advantages remain zero.
All-zero outcomes give zero future credit; historical targets still include
local penalties.

For the plotted ALFWorld contributions, let $\mu_x^d$ be the mean of $C^d$
over the restored union support and
$\sigma_x=s_x(C^{\mathrm H}+C^{\mathrm F})+10^{-6}$. For multiple supported rows,
\begin{equation}
A^d_{i,t}=\frac{C^d_{i,t}-\mu_{x_i}^d}{\sigma_{x_i}},
\qquad d\in\{\mathrm H,\mathrm F\},\qquad
A_{\mathrm{CC}}=A^{\mathrm H}+A^{\mathrm F}.
\label{eq:app-ccpo-component-decomposition}
\end{equation}
Both contributions are zero outside the union support. These additive contributions share a normalization scale and sum to $A_{\mathrm{CC}}$.

The optional episode channel uses
$e_{i,t}=R_i-\eta\mathbf1\{a_{i,t}\text{ invalid}\}$ over all restored task
rows, including contextually unsupported steps.
Hence longer trajectories contribute more rows, and local penalties can
produce different episode advantages within a trajectory. The channel is
added with unit weight after contextual normalization, without renormalizing.

\subsection{Policy-Update Integration}
\label{app:ccpo-policy}
The adapter broadcasts each detached step advantage to all generated response
tokens, including reasoning and executable actions; prompt and padding tokens
are excluded. The loss mask retains valid response tokens from steps with zero contextual credit. The dual-clipped policy objective~\citep{ye2020masteringcomplexcontrolmoba} follows Section~\ref{sec:method-optimization}.  Features and credit estimates receive
no gradients. Estimation uses float64 NumPy arithmetic, while the adapter
extracts and returns detached float32 features and token advantages.

For a response token $y$ with preceding context $c$, \texttt{low\_var\_kl} uses $\ell=\operatorname{clip}(\log\pi_{\mathrm{ref}}(y\mid c)-\log\pi_\theta(y\mid c),-20,20)$ and the penalty $\operatorname{clip}(e^\ell-\ell-1,-10,10)$. The penalty is aggregated with the response-token mask and added to the policy loss with coefficient $0.01$.

\medskip
\noindent\begin{minipage}{\linewidth}
\refstepcounter{ccpoalgorithm}\label{alg:ccpo-update}
\hrule\smallskip
\textbf{Algorithm \theccpoalgorithm\quad CCPO training iteration}
\smallskip\hrule\smallskip
\small
\begin{tabularx}{\linewidth}{@{}r@{\quad}X@{}}
1 & Collect complete rollouts with $\pi_{\mathrm{old}}$; gather metadata and reward targets.\\
2 & Extract frozen prompt features; retain unique steps and the input-row mapping.\\
3 & Encode prefix statistics; construct $\phi$ by Equation~\ref{eq:context-representation}.\\
4 & Select exact peers or task fallback; compute $\mathcal B[Y]$ and $V=\mathcal B[Z]$.\\
5 & Set $t^+\gets\min(t+K_{\mathrm F},T_i)$ and terminal values $V_{i,T_i}\gets R_i$.\\
6 & Set $C^{\mathrm H}_{i,t}\gets Y_{i,t}-\mathcal B_{i,t}[Y]$ and $C^{\mathrm F}_{i,t}\gets z_{x_i}[\Delta V]_{i,t}$, where $\Delta V_{i,t}=V_{i,t^+}-V_{i,t}$, on eligible steps.\\
7 & Zero-fill unavailable terms; restore rows; set $A_{\mathrm{CC}}\gets\mathcal N_x[C^{\mathrm H}+C^{\mathrm F}]$.\\
8 & Set $A\gets A_{\mathrm{CC}}+\alpha_{\mathrm{EP}}A_{\mathrm{EP}}$, $\alpha_{\mathrm{EP}}\in\{0,1\}$; pass this sum directly to the token adapter.\\
9 & Broadcast detached $A$ to response tokens; update the policy with PPO and KL.\\
\end{tabularx}
\smallskip\hrule
\end{minipage}

\section{Validation Rollout Examples}
\label{app:rollout-examples}

\begingroup
\setlength{\tabcolsep}{5pt}
\setlength{\LTpre}{\baselineskip}
\setlength{\LTpost}{\baselineskip}
\setlength{\LTcapwidth}{\linewidth}
\renewcommand{\arraystretch}{1.12}

\subsection{ALFWorld: Picking up a Soapbar}
\label{app:rollout-alfworld}
\begin{longtable}{@{}>{\raggedright\arraybackslash}p{0.16\linewidth}>{\raggedright\arraybackslash}p{\dimexpr0.84\linewidth-2\tabcolsep\relax}@{}}
\caption{\textbf{ALFWorld prompt and response excerpts at step 6 (Qwen2.5-1.5B-Instruct).}}\label{tab:rollout-alfworld}\\[\baselineskip]
\toprule
Component & Excerpt \\
\midrule
\endfirsthead
\toprule
Component & Excerpt (continued) \\
\midrule
\endhead
\bottomrule
\endfoot
\ccpoShadedRow\multicolumn{2}{@{}l@{}}{\strut\textbf{Prompt}}\\
Role & You are an expert agent operating in the ALFRED Embodied Environment. \\
\addlinespace[3pt]
Task & Your task is to: put two soapbar in garbagecan. \\
\addlinespace[3pt]
History & Prior to this step, you have already taken 5 step(s). Below are the most recent 2 observations and the corresponding actions you took: \\
\addlinespace[3pt]
History excerpt & [Observation 4: 'You arrive at cabinet 1. The cabinet 1 is closed.', Action 4: 'open cabinet 1'] \\
\addlinespace[3pt]
History excerpt & [Observation 5: 'You open the cabinet 1. The cabinet 1 is open. In it, you see nothing.', Action 5: 'go to sinkbasin 1'] \\
\addlinespace[3pt]
Current step & You are now at step 6 and your current observation is: You arrive at sinkbasin 1. On the sinkbasin 1, you see a soapbar 3, and a soapbar 2. \\
\addlinespace[3pt]
Actions & Your admissible actions of the current situation are: ['examine sinkbasin 1', [\ldots] 'take soapbar 2 from sinkbasin 1', 'take soapbar 3 from sinkbasin 1']. \\
\addlinespace[3pt]
Instructions & Now it's your turn to take an action. You should first reason step-by-step about the current situation. This reasoning process MUST be enclosed within \texttt{<}think\texttt{>} \texttt{<}/think\texttt{>} tags. Once you've finished your reasoning, you should choose an admissible action for current step and present it within \texttt{<}action\texttt{>} \texttt{<}/action\texttt{>} tags. \\
\addlinespace[3pt]
\ccpoShadedRow\multicolumn{2}{@{}l@{}}{\strut\textbf{Model response}}\\
Response & \texttt{<think>}\par The agent has arrived at the sinkbasin 1 and has two soapbars available: soapbar 2 and soapbar 3. The agent needs to put one of the soapbars in the garbagecan 1.\par\texttt{</think>}\par\smallskip \texttt{<action>take soapbar 2 from sinkbasin 1</action>} \\
\end{longtable}

\endgroup

\section{Statistical Analysis}
\label{app:stats}

\subsection{Relative Credit Preferences}
\label{app:stats:fig4a}

\paragraph{Data and filtering.}
\label{app:stats:scope}
For Figure~\ref{fig:ccpo-action-analysis}, we analyze ALFWorld training rollouts from iterations 101--150 of four
Qwen2.5-1.5B-Instruct runs: full CCPO and CCPO without the accumulated context summary,
both with $(K_{\mathrm H},K_{\mathrm F})=(2,2)$, together with the asymmetric $(4,0)$ and $(0,4)$ window variants of Table~\ref{tab:ablation-results-by-variant}. All use the $A_{\mathrm{CC}}$-only configuration.
Table~\ref{tab:stats-credit-samples} specifies the temporal windows and sample
counts. We retain successful trajectories, identified by a recorded episode
return greater than $0.5$, and interior steps satisfying $t\geq2$ and
$t_{\mathrm{last}}-t\geq2$, where $t$ is the zero-based logged step index
and $t_{\mathrm{last}}$ is the last logged step in the trajectory.
We apply this interior-step filter only to the analysis, consistently across all four runs. Padding duplicates are removed upstream; trajectories
with missing episode returns are excluded. Each run is analyzed using its
own rollouts, so comparisons across runs also reflect differences in the
sampled behavior.

\begin{table}[ht]
\caption{\textbf{Filtered samples for the credit-preference analysis.} Groups match
training iteration, task instance, and observation. $M$ counts categories with
at least 60 usable groups; each contrast requires both the category and other
actions in a group.}
\label{tab:stats-credit-samples}
\begin{center}
\small
\setlength{\tabcolsep}{4pt}
\begin{tabular}{@{}lcrrr@{}}
\toprule
Run & $(K_{\mathrm H},K_{\mathrm F})$ & Steps & Groups & $M$ \\
\midrule
CCPO (full)          & $(2,2)$ & 39,503 & 11,003 & 10 \\
w/o context summary & $(2,2)$ & 38,680 &  9,778 &  9 \\
Asymmetric (history) & $(4,0)$ & 40,972 &  9,914 & 10 \\
Asymmetric (future)  & $(0,4)$ & 49,736 & 13,235 & 11 \\
\bottomrule
\end{tabular}
\end{center}
\end{table}

\paragraph{Action categories.}
The action at step $t$ is classified from the environment response recorded
at step $t+1$, without using credit values. Fixed response-prefix rules
identify \emph{task operations} (clean, heat, or cool), \emph{take},
\emph{examine}, \emph{new location}, \emph{revisit}, \emph{open}, \emph{put},
\emph{toggle}, \emph{invalid}, \emph{close}, and \emph{other}.
Arrivals are classified as new locations or revisits using the locations
previously visited anywhere in the trajectory, including steps subsequently
removed by filtering. Invalid actions are identified by the environment's
``Nothing happens.'' response. Task completion may require revisiting a location.

\paragraph{Matched-group contrast.}
A comparison group $B$ contains steps from the same training iteration and
task instance with identical observation text. Only exact task-observation groups enter this analysis. For action category $c$, let $I_B(c)$ contain
steps of that category and $O_B(c)$ contain the other steps in $B$.
We retain groups with both sets nonempty and, for a step-level quantity $m_u$, define
\begin{equation}
\begin{split}
d_B(c;m)
&= \frac{1}{|I_B(c)|}\sum_{u\in I_B(c)}m_u
 - \frac{1}{|O_B(c)|}\sum_{u\in O_B(c)}m_u, \\
\Delta(c;m)
&= \frac{1}{|\mathcal U(c)|}
   \sum_{B\in\mathcal U(c)}d_B(c;m),
\end{split}
\label{eq:contrast}
\end{equation}
where $\mathcal U(c)$ is the set of usable groups. We apply this definition
to $A^{\mathrm H}$, $A^{\mathrm F}$, and $A_{\mathrm{CC}}$, using the additive
decomposition in Equation~\ref{eq:app-ccpo-component-decomposition}.
Groups receive equal weight regardless of their size.
A positive contrast indicates higher average group-relative credit for the category. Individual advantages may have either sign, and the contrast describes an association between action category and credit.

\paragraph{Bootstrap intervals and channel comparison.}
For each run, categories with fewer than 60 usable groups are excluded.
We draw 6,000 bootstrap samples of usable groups with replacement, using
random seed 0. Two-sided percentile intervals use quantile levels
$0.05/(2M)$ and $1-0.05/(2M)$, where $M$ is the number of eligible categories
in that run. Asterisks indicate intervals excluding zero. Each combination of run, analysis window, and measured quantity defines a separate Bonferroni correction family.

The channel comparison is paired within groups: we bootstrap
$d_B(c;A^{\mathrm F})-d_B(c;A^{\mathrm H})$ using the same sampled groups for
both terms. For new-location transitions in full CCPO, the mean paired
difference is $0.0372$ with adjusted interval $[0.0011,0.0727]$, based on
2,804 groups. The paired-contrast interval excludes zero; both channel-specific intervals include zero.

 The analysis is conditional on successful trajectories and
observations at which multiple action categories were sampled.

\subsection{Paired Action Timing During Evaluation}
\label{app:stats:fig4b}

\begin{figure}[!ht]
    \centering
    \includegraphics[width=\linewidth]{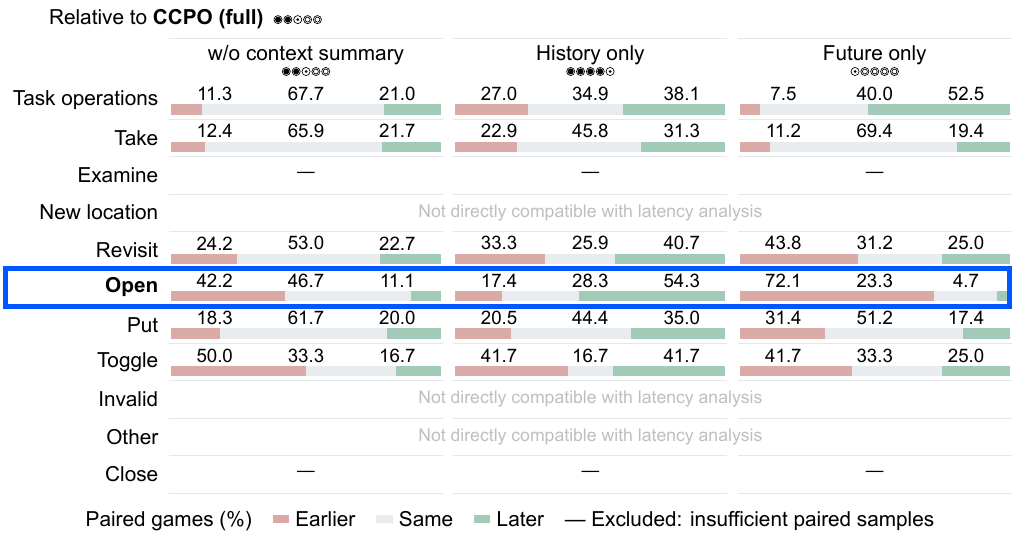}
    \caption{\textbf{Action timing on ALFWorld (Qwen2.5-1.5B-Instruct).} First occurrence at checkpoint 150, paired across the same 140 evaluation games against full CCPO. Percentages condition on both policies performing the action. Dashes: fewer than 12 pairs or unavailable results; gray rows are excluded. Context markers denote \ccpoAnchor~current, \ccpoHistory~history, and \ccpoFuture~future.}
    \label{fig:ccpo-action-timing}
\end{figure}

\paragraph{Evaluation population.}
For Figure~\ref{fig:ccpo-action-timing}, we use evaluation rollouts from checkpoints at training iteration 150, collected
following the main evaluation protocol. Each ablation is paired with full
CCPO on the same 140 ALFWorld tasks using their evaluation-game identifiers.
This analysis includes successful and failed episodes.

\paragraph{First occurrence.}
For game $g$ and category $c$, let $t_{g,c}$ denote the one-based step of its
first occurrence, identified by the response-based categories above.
The paired timing difference is
\begin{equation}
\Delta t_{g,c}
=t^{\mathrm{ablation}}_{g,c}-t^{\mathrm{full}}_{g,c}.
\end{equation}
Only games where both policies perform the action are eligible. The plotted
\emph{earlier}, \emph{same}, and \emph{later} percentages count pairs with
$\Delta t_{g,c}<0$, $=0$, and $>0$, respectively. Percentages are computed over eligible game pairs within each category. Games where
only one policy performs the action enter separate coverage counts and are
excluded from these percentages.

\paragraph{Displayed categories.}
Bars report descriptive proportions; dashes mark fewer than 12 eligible pairs
or unavailable results. New location is excluded: its first occurrence is fixed at step 1 under this labeling rule. Invalid and other are also excluded.

\subsection{Changes in Peer Contributions}
\label{app:stats:table3}

\paragraph{Paired representation comparison.}
For Table~\ref{tab:peer-reweighting}, we use full CCPO's Qwen2.5-1.5B-Instruct
ALFWorld training rollouts from iterations 101--150 to recompute
normalized peer weights with and without the accumulated context summary.
Rollouts, queries, recorded peer sets, trajectory exclusions, and return
targets are held fixed. The full representation has 1,573 dimensions;
the comparison retains its 1,536-dimensional hidden-state block and
renormalizes that block to unit length. The bandwidth rule is unchanged,
but its numerical value is recomputed from distances in each representation.
This comparison isolates the effect on weights for fixed rollouts; it is
separate from comparing independently trained ablation policies.

\paragraph{Population and statistic.}
We retain the recorded peer-pool selection, including task-only fallback
pools where used. This diagnostic covers successful and failed rollouts and retains boundary steps. It covers
99,424 queries and 1,678,986 query--peer pairs. For each pair, define the
absolute change in normalized weight, measured in percentage points, as
\begin{equation}
D_{uv}
=100\left|\bar w^{\mathrm{full}}_{u,v}
           -\bar w^{\mathrm{no\text{-}summary}}_{u,v}\right|.
\end{equation}
For a threshold $d$, the reported percentage is
\begin{equation}
P(d)=\frac{100}{|\mathcal Q|}
     \sum_{(u,v)\in\mathcal Q}\mathbf{1}\{D_{uv}>d\},
\end{equation}
where $\mathcal Q$ is the set of query--peer step pairs $(u,v)$. Pairs receive equal weight,
so queries with more peers contribute more terms. Thresholds of 0.25, 0.5,
1, 5, and 10 percentage points yield 62.1\%, 48.7\%, 33.8\%, 8.6\%, and
3.6\% of pairs, respectively. These are descriptive proportions of the
recorded pairs. These proportions quantify weight redistribution over fixed query-peer pairs.

\section{Failure Mode Comparisons}
\label{app:failure-comparisons}
We compare ALFWorld evaluation trajectories from Qwen2.5-1.5B-Instruct checkpoints at training step 150, with one sampled trajectory per method and task, temperature 0.4, and a 50-step limit. Tables show selected actions and observation excerpts; italicized rows summarize repeated actions.

\begingroup
\small
\setlength{\tabcolsep}{4pt}
\setlength{\LTpre}{6pt}
\setlength{\LTpost}{6pt}
\setlength{\LTcapwidth}{\linewidth}
\renewcommand{\arraystretch}{1.08}
\hyphenpenalty=10000
\newcommand{\ccpoTraceArm}[1]{\ccpoShadedRow\multicolumn{3}{@{}l@{}}{\strut\textbf{#1}}\\*}

\subsection{Repeated Visits Without Acquiring the Target}
\label{app:failure-revisits}
\ccpoHereFigure
    \centering
    \includegraphics[width=\linewidth]{ccpo/figures/main/revisit_ex.pdf}
    \ccpoHereCaption{\textbf{Evaluation trajectories for the navigation task in Table~\ref{tab:failure-revisits}.} Reproduced from Figure~\ref{fig:revisit-ex}. Each cell denotes one step; revisit counts track navigation to previously visited locations. CCPO uses $A_{\mathrm{CC}}$ only.}
    \label{fig:failure-revisits}
\ccpoEndHereFigure\par\medskip

\Needspace{8\baselineskip}
\begin{longtable}{@{}>{\raggedright\arraybackslash}p{0.10\linewidth}>{\raggedright\arraybackslash}p{0.39\linewidth}>{\raggedright\arraybackslash}p{\dimexpr0.51\linewidth-4\tabcolsep\relax}@{}}
\caption{\textbf{Repeated navigation versus target acquisition.} Task: ``put a toiletpaper in toiletpaperhanger.'' Revisit counts refer to navigation to previously visited locations.}\label{tab:failure-revisits}\\[4pt]
\toprule
Step & Action & Observation excerpt / summary \\
\midrule
\endfirsthead
\multicolumn{3}{@{}l}{\textit{Table \thetable\ continued}}\\
\toprule
Step & Action & Observation excerpt / summary \\
\midrule
\endhead
\bottomrule
\endfoot
\ccpoTraceArm{GiGPO: failed at 50 steps; 47 revisits}
3 & go to toiletpaperhanger~1 & On the toiletpaperhanger~1, you see nothing. \\
4--49 & \multicolumn{2}{p{\dimexpr0.90\linewidth-2\tabcolsep\relax}@{}}{\textit{46 navigation actions among countertop~1, toilet~1, and toiletpaperhanger~1; every returned observation repeats an earlier one.}} \\
\addlinespace[3pt]
\ccpoTraceArm{G2PO: failed at 50 steps; 42 revisits}
5 & open cabinet~2 & In it, you see a soapbottle~1. \\
8--15 & \multicolumn{2}{p{\dimexpr0.90\linewidth-2\tabcolsep\relax}@{}}{\textit{Two navigation cycles: countertop~1 $\to$ cabinet~2 $\to$ toilet~1 $\to$ toiletpaperhanger~1.}} \\
\addlinespace[3pt]
\ccpoTraceArm{HGPO: failed at 50 steps; 44 revisits}
4 & go to toiletpaperhanger~1 & On the toiletpaperhanger~1, you see nothing. \\
5--18 & \multicolumn{2}{p{\dimexpr0.90\linewidth-2\tabcolsep\relax}@{}}{\textit{14 navigation actions among toilet~1, countertop~1, and toiletpaperhanger~1; every returned observation repeats an earlier one.}} \\
\addlinespace[3pt]
\ccpoTraceArm{CCPO ($A_{\mathrm{CC}}$ only): success at step 7; 0 revisits}
3 & go to cabinet~1 & The cabinet~1 is closed. \\
4 & open cabinet~1 & In it, you see [\ldots] a toiletpaper~1. \\
5 & take toiletpaper~1 from cabinet~1 & You pick up the toiletpaper~1 from the cabinet~1. \\
6 & go to toiletpaperhanger~1 & On the toiletpaperhanger~1, you see nothing. \\
7 & move toiletpaper~1 to toiletpaperhanger~1 & You move the toiletpaper~1 to the toiletpaperhanger~1. \\
\end{longtable}

\subsection{Repeating a Completed Subgoal}
\label{app:failure-subgoal}
\ccpoHereFigure
    \centering
    \includegraphics[width=\linewidth]{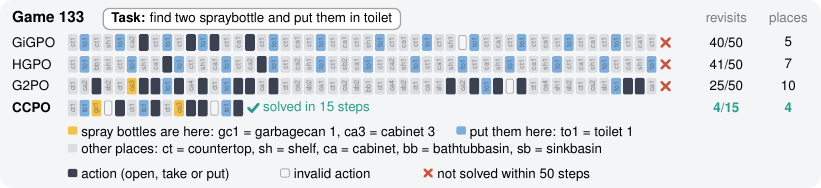}
    \ccpoHereCaption{\textbf{Evaluation trajectories for the two-object task in Table~\ref{tab:failure-subgoal}.} Each cell denotes one step; revisit counts track navigation to previously visited locations. CCPO uses $A_{\mathrm{CC}}$ only.}
    \label{fig:revisit-subgoal}
\ccpoEndHereFigure\par\medskip

\Needspace{8\baselineskip}
\begin{longtable}{@{}>{\raggedright\arraybackslash}p{0.10\linewidth}>{\raggedright\arraybackslash}p{0.39\linewidth}>{\raggedright\arraybackslash}p{\dimexpr0.51\linewidth-4\tabcolsep\relax}@{}}
\caption{\textbf{Repeated manipulation of one object versus placing two distinct objects.} Task: ``find two spraybottle and put them in toilet.''}\label{tab:failure-subgoal}\\[4pt]
\toprule
Step & Action & Observation excerpt / summary \\
\midrule
\endfirsthead
\multicolumn{3}{@{}l}{\textit{Table \thetable\ continued}}\\
\toprule
Step & Action & Observation excerpt / summary \\
\midrule
\endhead
\bottomrule
\endfoot
\ccpoTraceArm{G2PO: failed at 50 steps}
10 & move spraybottle~2 to toilet~1 & You move the spraybottle~2 to the toilet~1. \\
14 & go to toilet~1 & On the toilet~1, you see a cloth~3, and a spraybottle~2. \\
15 & take spraybottle~2 from toilet~1 & You pick up the spraybottle~2 from the toilet~1. \\
16 & move spraybottle~2 to toilet~1 & You move the spraybottle~2 to the toilet~1. \\
37--39, 48--49 & \multicolumn{2}{p{\dimexpr0.90\linewidth-2\tabcolsep\relax}@{}}{\textit{Two further take--replace cycles with spraybottle~2; step 38 is an invalid action.}} \\
\addlinespace[3pt]
\ccpoTraceArm{CCPO ($A_{\mathrm{CC}}$ only): success at step 15}
4 & ``1, and you have already taken '' & Nothing happens. \\
5 & take spraybottle~1 from garbagecan~1 & You pick up the spraybottle~1 from the garbagecan~1. \\
8 & move spraybottle~1 to toilet~1 & You move the spraybottle~1 to the toilet~1. \\
11 & open cabinet~3 & In it, you see a spraybottle~2. \\
12 & take spraybottle~2 from cabinet~3 & You pick up the spraybottle~2 from the cabinet~3. \\
13 & ``. now, you are at step 13, and'' & Nothing happens. \\
15 & move spraybottle~2 to toilet~1 & You move the spraybottle~2 to the toilet~1. \\
\addlinespace[3pt]
\ccpoTraceArm{GiGPO: failed at 50 steps}
11 & take soapbottle~2 from countertop~1 & You pick up the soapbottle~2 from the countertop~1. \\
13 & move soapbottle~2 to toilet~1 & You move the soapbottle~2 to the toilet~1. \\
\addlinespace[3pt]
\ccpoTraceArm{HGPO: failed at 50 steps}
1--50 & \multicolumn{2}{p{\dimexpr0.90\linewidth-2\tabcolsep\relax}@{}}{\textit{Navigation and cabinet opening; no object is picked up.}} \\
\end{longtable}
\endgroup

\section{Trajectory Lengths}
\label{app:trajectory-lengths}

\begin{figure}[!htbp]
    \centering
    \includegraphics[width=\linewidth]{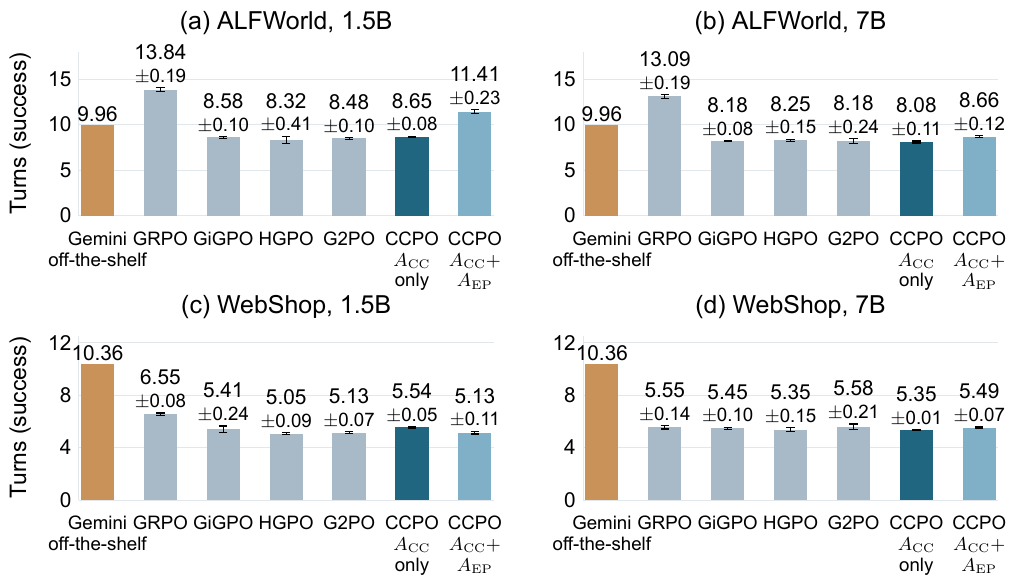}
    \caption{\textbf{Average steps among successful trajectories on ALFWorld and WebShop.} Panels show Qwen2.5-1.5B/7B-Instruct agents and a shared off-the-shelf Gemini reference. Error bars show sample standard deviations across three evaluation subsets; no deviation is reported for Gemini.}
    \label{fig:trajectory-lengths}
\end{figure}

Figure~\ref{fig:trajectory-lengths} reports average steps among successful trajectories. At 1.5B, the ALFWorld reference configuration ($A_{\mathrm{CC}}$) averages $8.65\pm0.08$ steps, compared with $5.13\pm0.11$ for the WebShop reference ($A_{\mathrm{CC}}+A_{\mathrm{EP}}$). WebShop's shorter successful trajectories motivate restricting the context-window variations to ALFWorld.

\ccpoHeadedFigure{Exact-Match Fallback and Peer Availability}{app:exact-match-fallback}

    \centering
    \includegraphics[width=\linewidth]{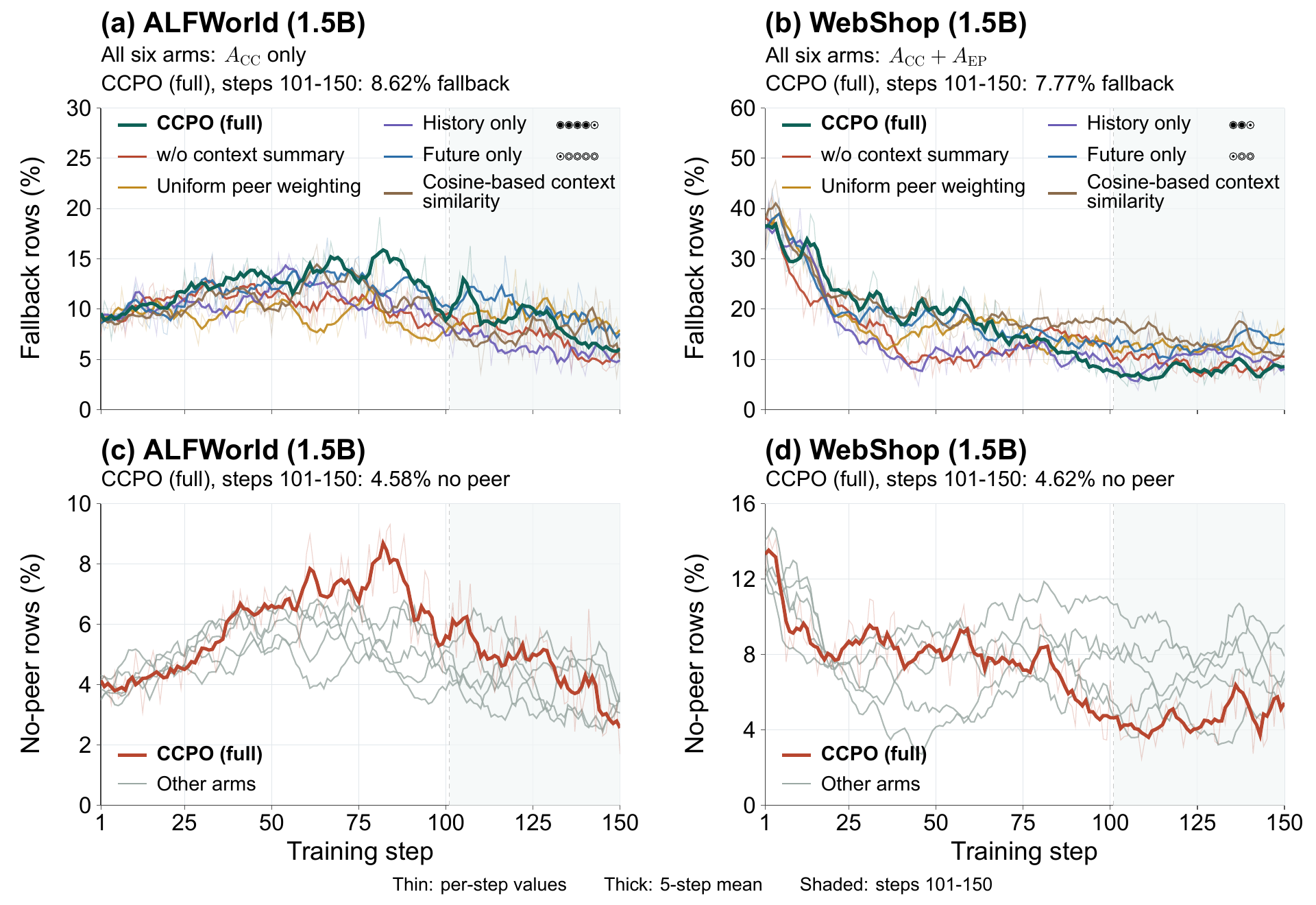}
    \ccpoHereCaption{\textbf{Exact-match fallback and peer availability during training.}
    Results use Qwen2.5-1.5B-Instruct on ALFWorld and WebShop.
    (a,b) Percentage of rows falling back from the exact anchor-observation key to a task-only key, comparing CCPO (full) with five ablations.
    (c,d) No-peer row percentages, computed as the singleton-bucket fraction divided by mean bucket size; CCPO (full) is highlighted and ablations are gray.
    Faint traces show raw per-step values, while smoothed curves use centered five-step moving averages.
    Shading marks steps 101--150; annotations report unweighted means of raw per-step values over this window.
    All ALFWorld arms use $A_{\mathrm{CC}}$ only, and all WebShop arms use $A_{\mathrm{CC}}+A_{\mathrm{EP}}$.
    Each arm represents one recorded training run.
    Vertical-axis ranges differ between benchmarks.}
    \label{fig:exact-match-fallback}
\ccpoEndHereFigure\par

\end{document}